\documentclass[10pt,journal,cspaper,compsoc]{IEEEtran}
\ifCLASSOPTIONcompsoc
\usepackage[nocompress]{cite}
\else
\usepackage{cite}
\fi

\usepackage[pdftex]{graphicx}
\DeclareGraphicsExtensions{.pdf,.jpg,.png,.eps}
\usepackage[pdfborder=false,colorlinks,bookmarksnumbered,bookmarksopen,linkcolor=black,citecolor=black,urlcolor=black]{hyperref}
\usepackage{graphicx}
\usepackage{mathrsfs,amsmath,amssymb} % define this before the line numbering.
\usepackage{array}
\usepackage{amsfonts}
\usepackage{footnote}
\usepackage{multicol}
\usepackage{booktabs}
\usepackage{threeparttable}
\usepackage{multirow}
\usepackage{color}
\usepackage{caption}
\def\ie{{\em i.e.}} %Jia: means `that is', usually used as an explanation of (\ie, xx)
\def\eg{{\em e.g.}} %Jia: means `for example', usually used to give some easy-to-understand examples (\eg, xx, xx and xx)
\def\etal{{\em et al.}} %Jia: used to enumerate authors when citing a paper. Only used when the paper has more than three authors.
\newcommand{\myPara}[1]{\vspace{.05in}\noindent\textbf{#1}} %Jia: a command that can be used to replace the itemize environment.
\newcommand{\bl}[1]{{\textbf{#1}}} % Jia: set text into bold, commonly used in: 1) dataset/model abbreviation, 2) best performance in tables
\newcommand{\ul}[1]{{\underline{#1}}} %Jia: set text into underline, commonly used to highlight the runner-up performance in tables
\newcommand{\mc}[1]{{\mathcal{#1}}} %Jia: cal fonts, commonly used to represent an instance such as image, superpixel, object
\newcommand{\mb}[1]{{\mathbb{#1}}} %Jia: bb fonts, used to represent a set, used together with \mc. For example, \mb{O}=\{\mc{O}_i\}_{i=1}^{N}
\usepackage{ragged2e}

\begin{document}
	
	\title{Complementary Segmentation of Primary Video Objects with Reversible Flows }
	
	\author{Jia~Li,~\IEEEmembership{Senior Member,~IEEE,}
		Junjie~Wu,	
		Anlin~Zheng,
		Yafei~Song,
		Yu~Zhang,
		and~Xiaowu~Chen,~\IEEEmembership{Senior~Member,~IEEE}
		
		\IEEEcompsocitemizethanks{
			\IEEEcompsocthanksitem J.~Li, J. Wu, A. Zheng and X. Chen are with State Key Laboratory of Virtual Reality Technology and Systems, School of Computer Science and Engineering, Beihang University, Beijing, 100191, China.\protect
			
			\IEEEcompsocthanksitem Y. Song is with National Engineering Laboratory for Video Technology, School of EE\texttt{\&}CS, Peking University, Beijing 100871, China. He is also with A.I. Labs, Alibaba Group, Beijing 100102, China. \protect
			
			\IEEEcompsocthanksitem Y. Zhang is with SenseTime Group Limited, 100084, China.  \protect
			
			\IEEEcompsocthanksitem A preliminary version of this work has been published in ICCV 2017~\cite{li2017primary}.\protect
			
%			\IEEEcompsocthanksitem J. Li is the corresponding author. URL: http://cvteam.net/ \protect
		}% <-this % stops an unwanted space
		%		\thanks{Manuscript received April 19, 2018; revised August 26, 2018.}
	}
	
	\markboth{~}%
	{Li \MakeLowercase{\textit{et al.}}: Bare Demo of IEEEtran.cls for Computer Society Journals}
	
	\IEEEtitleabstractindextext{
		\justifying
		\begin{abstract}
			Segmenting primary objects in a video is an important yet challenging problem in computer vision, as it exhibits various levels of foreground/background ambiguities. To reduce such ambiguities, we propose a novel formulation via exploiting foreground and background context as well as their complementary constraint. Under this formulation, a unified objective function is further defined to encode each cue. For implementation, we design a Complementary Segmentation Network (CSNet) with two separate branches, which can simultaneously encode the foreground and background information along with joint spatial constraints. The CSNet is trained on massive images with manually annotated salient objects in an end-to-end manner. By applying CSNet on each video frame, the spatial foreground and background maps can be initialized. To enforce temporal consistency effectively and efficiently, we divide each frame into superpixels and construct neighborhood reversible flow that reflects the most reliable temporal correspondences between superpixels in far-away frames. With such flow, the initialized foregroundness and backgroundness can be propagated along the temporal dimension so that primary video objects gradually pop-out and distractors are well suppressed. Extensive experimental results on three video datasets show that the proposed approach achieves impressive performance in comparisons with 18 state-of-the-art models.
		\end{abstract}
		
		}

	\maketitle
	
	\IEEEdisplaynontitleabstractindextext
	
	% \fi
	\IEEEpeerreviewmaketitle
	
	\IEEEraisesectionheading{\section{Introduction}\label{sec:introduction}}

	\IEEEPARstart{S}{egmenting} primary objects aims to delineate the physical boundaries of the most perceptually salient objects in an image or video. By perceptual saliency, it means that the objects should be visually salient in image space while present in most of the video frames. This is an useful assumption that works under various unconstrained settings, thus benefiting many computer vision applications such as action recognition, object class learning, video summarization, video editing and content-based video retrieval.
	
	\begin{figure}[!t]
		\centering {
			\includegraphics[width=1.0\columnwidth]{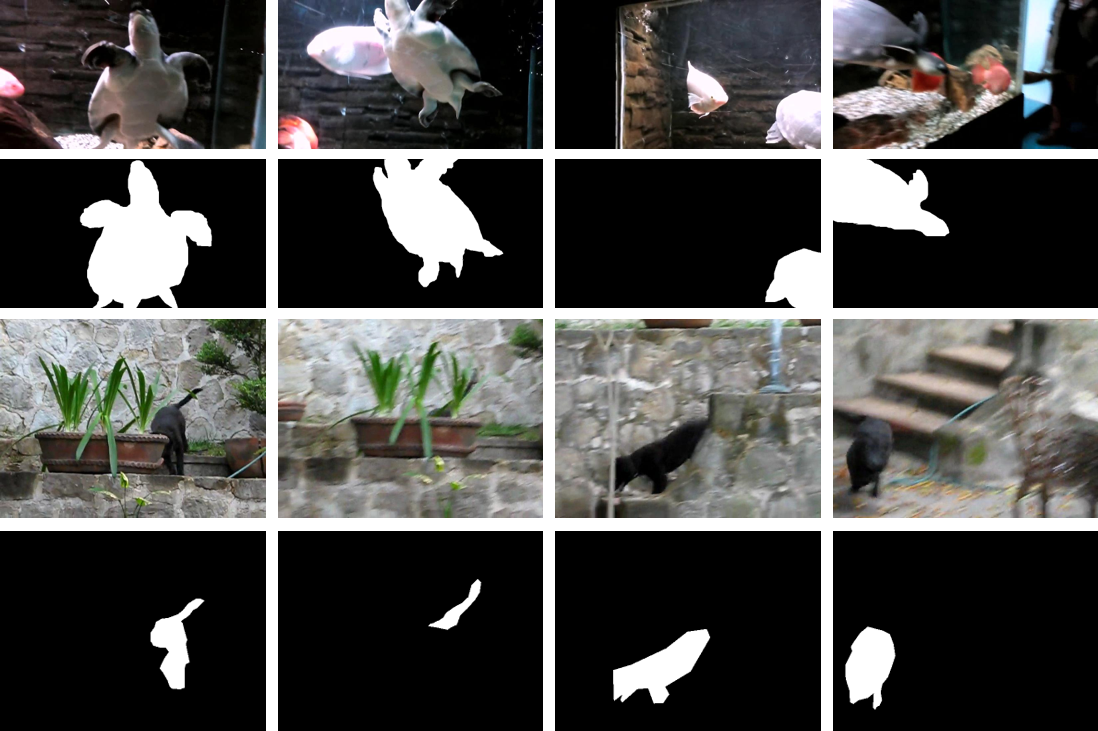}}  %模板是2.5in  %原来的ICCV是1.0\columnwidth
		\caption{Primary objects may co-occur with or be occluded by various distractors. They may not always be the most salient ones in each separate frame but can consistently pop-out in most video frames (frames and masks are taken from the datasets \bl{VOS}\cite{VOSDataset} \bl{Youtube-Objects}~\cite{prest2012youtubeobjects}, respectively).}
		\label{Fig:primary}
	\end{figure}
	
	Despite impressive performance in recent years~\cite{JiangWYWZL13,han2016co,lee2016eld,li2015mdf,wang2015legs,zhao2015mcdl}, primary object segmentation remains a challenging task since in real world images there exist various levels of ambiguities in determining whether a pixel belongs to the foreground or background. The ambiguities are more serious in video frames due to some video attributes representing specific situations, such as fast-motion, occlusion, appearance change and cluttered background~\cite{Perazzi2016}. Specially, these attributes are not exclusive, thus a sequence can be annotated with multiple attributes.  As shown in Fig.~\ref{Fig:primary}, due to the camera and/or object motion, the primary objects may suffer motion blur (\eg, the last dog frame), occlusion (\eg, the second dog frame) and even out-of-view (\eg, the last two turtle frames). Moreover, the primary objects may co-occur with various distractors in different frames (\eg, the turtle video frames), making them difficult to consistently pop-out throughout the whole video.

	%In recent years, image-based primary object segmentation has achieved impressive performance since powerful models can be directly trained on large image datasets by using Random Forest \cite{JiangWYWZL13}, Multi-instance Learning~\cite{han2016co}, Stacked Autoencoders~\cite{han2015drr} and Deep Neural Networks~\cite{lee2016eld,li2015mdf,wang2015legs,zhao2015mcdl}. In contrast, segmenting primary video objects remains a challenging task since the amount of video data with pixel-level annotations is much less than that of images, which may prevent the end-to-end training of a sophiscated spatio-temporal model.
	
	%Even if the end-to-end training is not required, directly migrating/extending image based algorithms to video may bring inconsistent segmentation although it is nontrivial;
	% 2) directly migrating/extending image based algorithms to video may bring inconsistent segmentation although it is nontrivial;
	%   Moreover, due to the camera and/or object motion, the primary objects may co-occur with or be occluded by various distractors in different frames (see Fig.~\ref{Fig:primary}), making them difficult to consistently pop-out throughout the whole video.

	\begin{figure*}[!t]
		\centering {
			\includegraphics[width=1.00\textwidth]{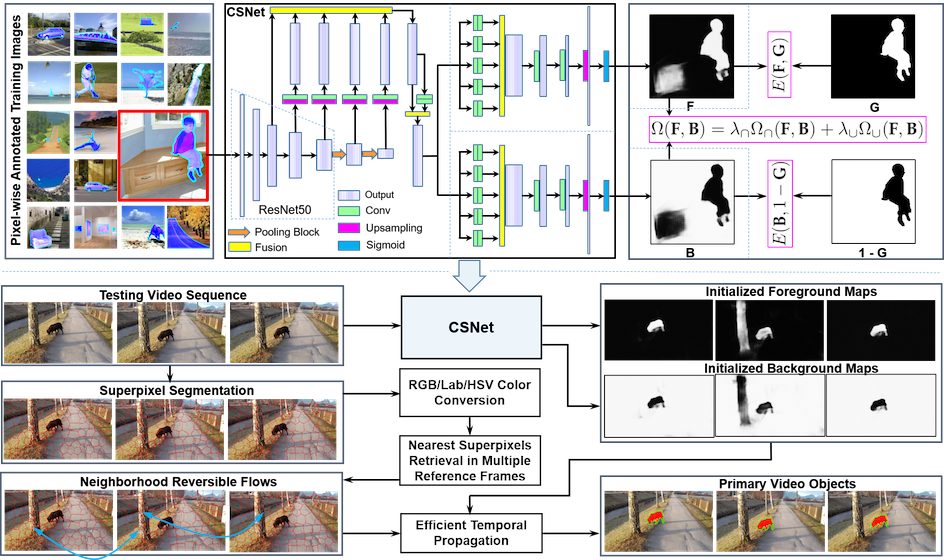}}
		\caption{Framework of the proposed approach. The framework consists of two major modules. The spatial module trains CSNet to simultaneously initialize the foreground and background maps of each frame. This module operates on GPU to provide pixel-wise predictions for each frame. The temporal module constructs neighborhood reversible flow so as to propagate foregroundness and backgroundness along the most reliable inter-frame correspondences. This module operates on superpixels for efficient temporal propagation. Note that $E(\cdot)$ is the cross-entropy loss that enforce $F \rightarrow G$ and $B \rightarrow 1-G$. The proposed complementary loss $\Omega(F,B)$ contains intersection loss $\Omega_{\cap}(F,B)$ and union loss $\Omega_{\cup}(F,B)$ for a complementary constraint. $F$, $B$ and $G$ are foreground, background and groundtruth, respectively. $\lambda_{\cap}$ and $\lambda_{\cup}$ are corresponding weights. Moreover, more details about CSNet are shown in \ref{sec:Initialization CNN}. }
		\label{Fig:framework}
	\end{figure*}

	To address these issues, there exist three major types of models which can be roughly categorized into interactive, weakly-supervised and fully-automatic ones. Interactive models require manually annotated primary objects in the first frame or several selected frames before the automatic segmentation~\cite{maerki2016bilateral,ramakanth2014seamseg,seguin2016instance}, while weakly-supervised ones often assume that the semantic tags of primary video objects are known before segmentation so that external cues like object detections can be used~\cite{tsai2016coseg,zhang2015sosd}. However, the requirement of interaction or semantic tags prevents their usage in processing large-scale video data~\cite{zhang2013satd}.
	
	Beyond the two kinds of models, fully-automatic models aim to directly segment primary objects in a single video~\cite{jang2016primary},~\cite{papazoglou2013fst},~\cite{zhang2013satd,xiao2016track} or co-segment the primary objects shared by a collection of videos~\cite{chiu2013coseg},~\cite{Fu15Video},~\cite{zhang2014video} without any prior information about objects.  Although CNNs have achieved impressive progress in object segmentation, insufficient video data with pixel-level annotations may prevent the end-to-end training of a complex spatiotemporal model.

	In view of remarkable performance in image-based primary object segmentation, an easy way is to extend the image-based models to videos by considering spatial attributes and the additional temporal cues of primary video objects~\cite{yu2015efficient,jang2016primary,wang2018saliency}. Such spatiotemporal attributes like attractive appearance, better objectness, distinctive motion from its surroundings and frequent occurrence in the whole video, mainly focus on foreground features and have attracted much attention from most models~\cite{cheng2015global,jiang2013salientCvpr,jiang2013salient}. While actually background is symbiotic with foreground and contains much connotative information. Thus some models pay more attention to background cues, such as boundary connectivity~\cite{cheng2015global,zhu2014saliency}, surroundings~\cite{zhang2013saliency}, even including complex dynamic background modeling~\cite{ge2016dynamic}. Naturally it leads to several models~\cite{koh2017primary,jang2016primary} that consider both foreground and background cues to assist foregroundness segmentation. However, there exist two issues. On one hand, sometimes the complexity of primary objects renders these attributes insufficient (\eg, distractors share common visual attributes with targets), then these models may fail on certain videos in which the assumptions may not hold. On the other hand, these models either ignore foreground/background or only utilize one to facilitate the other, which may miss some important cues and result in more ambiguities between foreground and background.
	
	Moreover, temporal coherence is an important issue for primary video object segmentation, and directly applying image based algorithms to videos is vulnerable to inconsistent segmentation. To reduce such inconsistency, costly processing steps are usually adopted, such as object/trajectory tracking and sophisticated energy optimization models~\cite{wang2018saliency,tsai2016video,koh2017primary,jang2016primary}. Particularly, pixel-wise optical flows are widely used to propagate information between adjacent frames. Unfortunately, optical flows are often inaccurate in case of sudden motion changes or occlusions, by which errors may be accumulated along time. Moreover, the correspondences in adjacent temporal windows may prevent long-term information being propagated more effectively.

	Considering all these issues, this paper proposes a novel approach that effectively models the complementary nature of foreground/background in primary video object segmentation, and efficiently propagates information temporally within neighborhood reversible flow (NRF). Firstly, the problem of primary object segmentation is formulated into a novel objective function that explicitly considers foreground and background cues as well as their complementary relationships. In order to optimize the function and obtain the foregroundness and backgroundness prediction, a Complementary Segmentation Network (CSNet) with multi-scale feature fusion and foreground/background branching is proposed. Then, to enhance the temporal consistency of initial predictions, NRF is further proposed to establish reliable, non-local inter-frame correspondences. These two techniques constitute into the spatial and temporal modules of the proposed framework, as shown in Fig.~\ref{Fig:framework}.~
	%
	%Two steps: first, starting from segmentation essence, obtain initial predictions by formulating the problem of primary object segmentation from a complementary perspective based on foreground and background prior and optimizing it based on a Complementary Convolutional Neural Networks (CCNN); second, aiming to smooth the temporal domain, propagate reliable inter-frame information using NRF and yield consistent segmentation.
	%The framework of the proposed approach is shown in Fig.~\ref{Fig:framework}, which mainly consists of two modules.
	
	In the spatial module, CSNet is trained on massive annotated images as an optimizer of the proposed complementary objective so as to simultaneously handle two complementary tasks, \ie, foregroundness and backgroundness estimation, with two separate branches. By using CSNet, we can obtain the initialized foreground and background maps on each individual frame. To efficiently and accurately propagate such spatial predictions between far-away frames, we further divide each frame into a set of superpixels and construct neighborhood reversible flow so as to depict the most reliable temporal correspondences between superpixels in different frames. Within such flow, the initialized spatial foregroundness and backgroundness are efficiently propagated along the temporal dimension by solving a quadratic programming problem that has analytic solution. In this manner, primary objects can efficiently pop-out and distractors can be further suppressed. Extensive experiments on three video datasets show that the proposed approach acts efficiently and achieves impressive performances compared with 18 state-of-the-art models (7 image-based \& non-deep, 6 image-based \& deep, 5 video-based).
	%\begin{CJK*}{GBK}{song}
	%(\bl{下面关于和会议论文的比较应该再扩展下，包括contributions 应该再仔细写下})
	%\end{CJK*}
	This paper builds upon and extends our previous work in~\cite{li2017primary} with further discussion of the algorithm, analysis and expanded evaluations. We further formulate the segmentation problem into a new objective function based on the constraint relationship between foreground and background and optimize it using a new complementary deep networks.
	\newline
	\indent The main contributions of this paper include:\\
	1)~we formulate the problem of primary object segmentation into a novel objective function based on the relationship between foreground and background, and incorporate the objective optimization problem into end-to-end CNNs. In this manner, two dual tasks of foreground and background segmentation can be simultaneously addressed and primary video objects can be segmented from complementary cues.\\
	~~~~2)~we construct neighborhood reversible flow between superpixels which effectively propagates foreground and background cues along the most reliable inter-frame correspondences and leads to more temporally consistent results.\\
	~~~~3)~Based on the proposed method, we achieve impressive performance compared with 18 image-based and video-based existing models, achieving state-of-the-art results.
	
	%     1)~we design a novel lost function based on foreground and background prior that formulates the problem of primary object segmentation from a complementary perspective;\\
	%    2)~we propose a simple yet effective framework for efficient and accurate primary video object segmentation;\\
	%    3)~we incorporate the optimization problem of cost function into end-to-end CNNs, by training specific CNNs to simultaneously address two dual tasks of foreground and background segmentation so that primary video objects can be detected from complementary perspectives;\\
	%    4)~we construct neighborhood reversible flow between superpixels that can effectively propagate foregroundness and backgroundness along the most reliable inter-frame correspondences.
	In the rest of this paper, we first conduct a brief review of previous studies on primary/salient object segmentation in Section 2. Then, we present the technical details of the proposed spatial initialization module in Section 3 and temporal refinement module in Section 4. Experimental results are shown in Section 5. At last, we conclude with a discussion in Section 6.

	\section{Related Work}\label{sec:Related Work}
	
	%primary video object segmentation, semantic video segmentation, image segmentation.\\
	%%主要是三部分，primary video object segmentation， semantic video segmentation， image segmentation。对image分割部分的描述可以简略些，只挑代表性的工作说。可以参考实验中所对比的那些方法进行做。
	A great performance of primary video object segmentation is contributed by good performance of each frame.
	In this section, we give a brief overview of recent works in salient object segmentation in images and primary/semantic object segmentation in videos.
	
	\subsection{Salient Object Segmentation in Images}\label{sec:Related Work image}
	
	%Salient object segmentation is commonly interpreted as a process of two stages:1)detecting the most salient object and 2)segmenting the accurate region of that object.
	Salient object segmentation in images is a research area that has been greatly developed in the past twenty years in particular since 2007~\cite{liu2007region}.
	%Based on different mechanisms, salient object segmentation models can generally be divided into two categories. The first are rapid, bottom-up, and data-driven approaches without semantic information, while the second are slow, top-down, and task-driven methods with semantic information.
	
	%objectness~\cite{jiang2013salient,zhang2013saliency},, including the objectness which leverages object proposals by measuring the likelihood of there being a complete object around a pixel or region~\cite{alexe2012measuring}
	%the widely used contrast prior believes that the salient regions present high contrast over background in certaincontext~\cite{goferman2012context,liu2011learning,jiang2013salientCvpr,perazzi2012saliency,hou2007saliency},
	
	Early approaches treat saliency object segmentation as an unsupervised problem and focus on low-level and mid-level cues, like contrast~\cite{cheng2015global,perazzi2012saliency}, focusness~\cite{Jiang_2013_ICCV}, spatial property~\cite{goferman2012context,wei2012geodesic}, spectral information~\cite{schauerte2012quaternion}, objectness~\cite{zhang2013saliency}, etc. Most of the cues build upon foreground priors. For example, the widely used contrast prior believes that the salient regions present high contrast over background in certain context~\cite{goferman2012context,liu2011learning}, and the focusness prior considers that a salient object is often photographed in focus to attract more attention. From the opposite perspective, background prior is first proposed by Wei \etal ~\cite{wei2012geodesic}, who assume the image boundaries are mostly background and build a saliency detection model based two background priors, \ie, boundary and connectivity. After that, some approaches~\cite{YangZLRY13Manifold,zhu2014saliency,Qin2015BSCA,Zhang2015MB} successively appear.
	%Zhu \etal ~\cite{zhu2014saliency} utilize the geodesic distance and take the spatial layout of image patches into consideration for a more robust boundary measurement.
	%In~\cite{Qin2015BSCA}, Qin \etal apply the background prior to compute a spatial distance map as well as a global color distinction map, and integrate them into a course background-based map for initialization.
	%Based on the low-level features, objectness as a mid-level cues, measuring the likelihood of there being a complete object around a pixel or region~\cite{alexe2012measuring}, is also used to faciliate salient object segmentation by leveraging object proposals~\cite{jiang2013salient,zhang2013saliency}.
	% The measurement is calculated by fusing hybrid low level features such as multi-scale saliency, color contrast, edge density and superpixels straddling.
	%three foreground visual cues namely uniqueness, focusness and objectness
	%Based on the independent cues, several approaches~\cite{jiang2013salient,zeng2018unsupervised} integrate some of them into an optimization formulation for a joint foreground/background guidence. In \cite{jiang2013salient}, three foreground visual cues namely uniqueness, focusness and objectness are used to estimate the final saliency map. In \cite{zeng2018unsupervised}, saliency segmentation problem is formulated as a non-cooperative game in which image regions choose to be background or foreground based on a payoff function.
	Unfortunately, these methods usually require a prior hypothesis about salient objects and their performance heavily depend on the prior reliability. Besides, the methods that only use purely low-level/mid-level cues are difficult to detect salient objects in complex scenes due to unawareness of image content.
	
	Recently, learning based methods, especially deep networks methods~(\ie, CNN-based models and FCN-based models), attract much attention because of the ability to extract the high-level semantic information~\cite{li2016deepsaliency,lee2016eld,wang2015legs}. In~\cite{wang2015legs}, two neural networks DNN-L and DNN-G are proposed to respectively extract local features and conduct a global search for generating the final saliency map. In~\cite{li2015mdf}, Li and Yu introduce a neural network with fully connected layers to regress the saliency degree of each superpixel by extracting multiscale CNN features. While these CNN-based models with fully connected layers that operate at the patch-level may result in blurry saliency maps, especially near the boundary of salient objects, thus in~\cite{long2015fully}, fully convolutional networks considering pixel-level operations is applied for salient object segmentation. After that, various FCN-based salient object segmentation approaches are explored~\cite{li2016deep,liu2016dhsnet,kuen2016recurrent} and obtain impressive performance.
	%For instance, Li \etal ~\cite{li2016deep} design a end-to-end deep contrast network with two complementary branches: a pixel-level fully convolutional stream and a segment-wise spatial pooling stream, and the first stream is a multi-scale fully convolutional network to infer a pixel-level saliency map directly from the raw input image.
	
	However, most of the methods focus on independent foregroud or background features and only several models~\cite{yu2010automatic,tian2015learning} pay attention to both of them. While to the best of our knowledge, few models explicitly model the constraint relationship between them although it may be very helpful in complex scenes. Therefore, in this work, we simultaneously consider foreground and background cues as well as their complementary relationships and optimize their joint objective by using the powerful learning ability of deep networks.
	%
	% cluttered background in complex scenes may contain more useful information than foreground targets. actually foreground and background should depend on and constraint each other
	\subsection{Primary/Semantic Object Segmentation in Videos}\label{sec:Related Work video}
	Different from salient object segmentation in images, primary video object segmentation face more challenges and criteria (\eg, spatiotemporal consistency) due to the additional temporal attributes. % such as motion, recurrence.
	%Motion information~(\eg, motion vectors and feature point trajectories) is usually used in spatiotemporal domain to facilitate primary video object segmentation~\cite{gao2008discriminant,mahadevan2010spatiotemporal,fang2014video,liu2014superpixel,rahtu2010segmenting,seo2009static,wang2015sag,ochs2014moseg}.
	
	Motion information~(\eg, motion vectors, feature point trajectories and optical flow) is usually used in spatiotemporal domain to facilitate primary/semantic video object segmentation and enhance the spatiotemporal consistency of segmentation results ~\cite{liu2014superpixel,wang2015sag,ochs2014moseg}.
	%Some early methods~\cite{gao2008discriminant,seo2009static,mahadevan2010spatiotemporal} simply incorporate motion cues into image object segmentation model as a feature or calculate the motion saliency map to refine each frame segmentation. While these methods often ignore the spatiotemporal consistency in video sequences.
	For example, Papazoglou and Ferrari~\cite{papazoglou2013fst} first initialize foreground maps with motion information and then refine them in the spatiotemporal domain so as to enhance the smoothness of foreground objects. Zhang~\etal~\cite{zhang2013satd} use optical flow to track the evolution of object shape and present a layered Directed Acyclic Graph based framework for primary video object segmentation.
	In a further step, Tsai~\etal~\cite{tsai2016video} utilize a multi-level spatial-temporal graphical model
	with the use of optical flow and supervoxels to jointly optimize segmentation and optical flow in an iterative scheme. The re-estimated optical flow (\ie, object flow) is used to maintain object boundaries and temporal consistency.
	Nevertheless, there still exist several issues. Firstly, some models~\cite{li2013segtrackv2,wang2015sag,wang2015GF} are built upon certain assumptions, for instance foreground objects should move differently from its surroundings in a good fraction of the video or should be spatially dense and change smoothly across frames in shapes and locations, which may fail on certain videos that contain complex scenarios in which assumptions may not hold. Secondly, the pixel-wise optical flow are usually computed between adjacent frames since their similarity can offer more accurate flow estimation, while it is disadvantageous to obtain more valuable inter-frame (\eg, two far-away frames) cues since adjacent frames may not offer useful cues due to occlusion, blur and out-of-view, etc.

	Recently, a number of approaches attempt to address video object segmentation via deep neural networks. While due to lacking sufficient video data with per-frame pixel-level annotations, most of them exploit temporal information over image segmentation approaches for video segmentation. One popular thought is to calculate a kind of correspondence flow and propagate it in inter-frames~\cite{nilsson2016semantic,gadde2017semantic,zhu2017deep}. In~\cite{nilsson2016semantic} based on optical flow, a Spatio-Temporal Transformer GRU is proposed to temporally propagate labeling information between adjacent frames for semantic video segmentation. In~\cite{zhu2017deep} a deep feature flow is presented to propagate deep feature maps from key frames to other frames, which is jointly trained with video recognition tasks. Although these methods are helpful for transfering image-based segmentation networks to videos, the propagation flows are still limited by adjacent frames or training complexity.
	
	Therefore in our work, we enhance inter-frame consistency by constructing neighborhood reversible flow(NRF) instead of optical flow to efficiently and accurately propagate the initialized predictions between adjacent key frames, which is simple but effective for popping out the consistent and primary object in the whole video.
	
	%  Clockwork . Some CNN-based methods~\cite{perazzi2017learning,caelles2017one} are developed to find objects in videos by combining offline and online training processes on static images and fine-tuning on the first frame in the sequence. Although outstanding performance has been achieved, the segmentation results are not guaranteed to be smooth in the temporal domain.
	%Though our method also learns salient features from static images, our training process is more like a deep optimization strategy starting from a cost function. Moreover, we use NRF to enhance temporal smooth and yield more consistent segmentation results.
	
	\section{Initialization with Complementary CNNs}\label{sec:Initialization}
	
	In this section, starting from the complementary peculiarity of foreground and background, we reformulate the problem of primary video object segmentation into a new objective function. Then we design complementary CNNs to conduct deep optimization of the objective function and yield the initial foreground and background estimation.

	\subsection{Problem Formulation}\label{sec:Initialization Formulation}

	Typically, a frame $\mc{I}$ consists of the foreground area $\mc{F}$ and the background area $\mc{B}$ with $\mc{F}\cap\mc{B}=\O$ and $\mc{F}\cup\mc{B}=\mc{I}$, \ie, the foreground and background should be complementary in image space. Considering that foreground objects and background distractors usually have different visual characteristics (\eg, clear versus fuzzy edges, large versus small sizes, high versus low objectness), we can attack the problem of primary object segmentation at the frame $\mc{I}$ from a complementary perspective, estimating foreground and background maps, respectively. In this manner, the intrinsic characteristics of foreground and background regions can be better captured by two models with different focuses. Keeping this in mind, we propose the following formulation to explicitly consider foreground and background cues
	%In view of the symbiosis and constraint relationship between foreground and background, formally the two models can be learned by solving a cost function
	\begin{equation}\label{eq:origCCNN}
	\begin{split}
	\min_{\mb{W}_F,\mb{W}_B}& \mc{L}(\bl{F}, \bl{B}, \bl{G})+\Omega(\bl{F}, \bl{B}),\\
	\text{s.t.}&~\phi_F(\mc{I};\mb{W}_F)=\bl{F},~\bl{F}(p)\in\{0,1\},\forall~p\in{}\mc{I}\\
	&~\phi_B(\mc{I};\mb{W}_B)=\bl{B},~\bl{B}(p)\in\{0,1\},\forall~p\in{}\mc{I},
	\end{split}
	\end{equation}
	where $\bl{F}$ and $\bl{B}$ are two binary matrices representing $\mc{F}$ and $\mc{B}$. $\bl{G}$ is the ground-truth map that equals 1 for a foreground pixel and 0 for a background pixel. $\mb{W}_F$ and $\mb{W}_B$ are two sets of parameters for the foreground and background prediction models $\phi_F$ and $\phi_B$. For the sake of simplifications, the values of $\bl{F}$ and $\bl{B}$ are assumed to be in the range [0,1]. The first term $\mathcal L(\bl{F}, \bl{B},\bl{G})$ is the empirical loss defined as
	\begin{equation}\label{eq:lossCCNN}
	\mc{L}(\bl{F}, \bl{B}, \bl{G})=E(\bl{F}, \bl{G}) + E(\bl{B},1-\bl{G}),
	\end{equation}
	where $E(\cdot)$ is the cross-entropy loss that enforce $\bl{F}\rightarrow{}\bl{G}$ and $\bl{B}\rightarrow{}1-\bl{G}$.
	Ideally, salient objects and background regions can be perfectly detected by minimizing these two losses. However, errors always exist even when two extremely complex models are used. In this case, conflicts and unlabeled areas may arise in the predicted maps (\eg, both $\bl{F}$ and $\bl{G}$ equals 1 or 0 at the same location).
	%, see Fig.~\ref{Fig:complementary}(b)-(e)).
	
	To reduce such errors, we refer to the constraint relationship $\mc{F}\cap\mc{B}=\O$ and $\mc{F}\cup\mc{B}=\mc{I}$ and incorporate the complementary loss $\Omega(\bl{F}, \bl{B})$:
	\begin{equation}\label{eq:priorCCNN}
	%\Omega(\bl{F}, \bl{B}) = \lambda_\cap\Omega_\cap(\bl{F}, \bl{B}) + \lambda_\cup \Omega_\cup(\bl{F}, %\bl{B}),
	\Omega(\bl{F}, \bl{B}) = \lambda_\cap\Omega_\cap(\bl{F}, \bl{B}) + \lambda_\cup \Omega_\cup(\bl{F}, \bl{B}),
	\end{equation}
	where $\Omega_\cap(\cdot)$ and $\Omega_\cup(\cdot)$ are two losses with non-negative weights $\lambda_\cap$ and $\lambda_\cup$ to encode the constraint $\mc{F}\cap\mc{B}=\O$ and $\mc{F}\cup\mc{B}=\mc{I}$, respectively. Here, $\lambda_\cap$ and $\lambda_\cup$ are both set as 0.4. The intersection loss term $\Omega_\cap(\cdot)$ tries to minimize the conflicts between $\bl{F}$ and $\bl{B}$:
	\begin{equation}\label{eq:intersactionLoss}
	\Omega_\cap(\bl{F}, \bl{B})=\frac{1}{\|\mc{I}\|}\sum_{p\in\mc{I}}\left(\bl{F}(p)\cdot{}\bl{B}(p)\right)^{\sigma_\cap},
	\end{equation}
	where $\|\mc{I}\|$ indicates the number of pixels in the image $\mc{I}$ and $p$ is a pixel with predicted foregroundness $\bl{F}(p)$ and backgroundness $\bl{B}(p)$. $\sigma_\cap$ is a positive weight to control the penalty of conflicts. The minimum value of \eqref{eq:intersactionLoss} will be reached when $\bl{F}(p)\cdot\bl{B}(p)=0$, implying that at least one map has zero prediction at every location.
	
	%no hole in F+B
	Similarly, the union loss term $\Omega_\cup(\cdot)$ tries to maximize the complementary degree between $\bl{F}$ and $\bl{B}$:
	\begin{equation}\label{eq:unionLoss}
	\Omega_\cup(\bl{F}, \bl{B})=\frac{1}{\|\mc{I}\|}\sum_{p\in\mc{I}}\left(\bl{F}(p)+\bl{B}(p)-1\right)^{\sigma_\cup}.
	\end{equation}
	We can see that the minimum complementary loss can be reached when $\bl{F}(p)+\bl{B}(p)=1$ (\ie, perfect complementary predictions). The parameter $\sigma_\cup$ is a positive weight to control the penalty of non-complementary predictions.

	\subsection{Deep Optimization with Complementary CNNs}\label{sec:Initialization CNN}

	Given the empirical loss~\eqref{eq:lossCCNN} and the complementary loss~\eqref{eq:priorCCNN}, we can derive two models $\phi_F(\cdot)$ and $\phi_B(\cdot)$ for per-frame initialization of the foreground and background maps by solving the optimization problem of objective function~\eqref{eq:origCCNN}. Toward this end, we need to first determine the form of the models and the algorithm for optimizing their parameters. Considering the impressive capability of convolutional neural network (CNN), we propose to solve the optimization problem in a deep learning paradigm.
	
\begin{figure}[!t]
		\centering {
			\includegraphics[width=1.0\columnwidth]{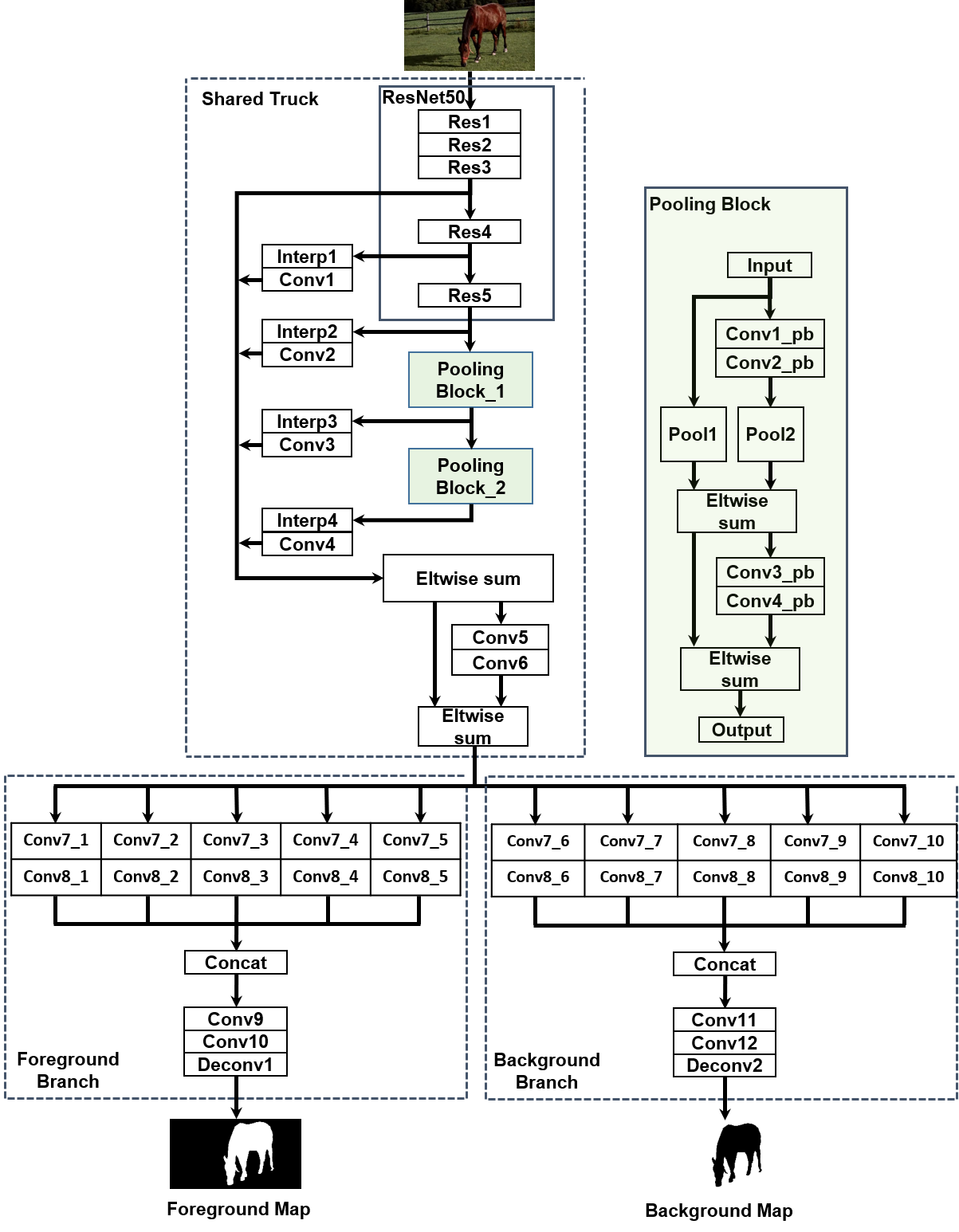}} %width=1.0
		%\caption{Architecture of CCNN. Here '\texttt{CONV} ($3\times{}3/2$)' indicates a convolutional layer with $3\times3$ kernels and dilation of 2.}
		\caption{Architecture of the proposed CSNet. Note that layer \texttt{Res1} and \texttt{Res2/3/4/5} correspond to layer \texttt{conv1} and \texttt{conv2\_x/3\_x/4\_x/5\_x} in~\cite{he2016deep}, respectively. More details are shown in Table \ref{tab:CSNet architectures}.}
		\label{Fig:CCNN}
	\end{figure}

	The architecture of the proposed CNN can be found in Fig.~\ref{Fig:CCNN}, which starts from a shared trunk and ends up with two separate branches, \ie, foreground branch and background branch. Main configurations and details are shown in Table~\ref{tab:CSNet architectures}. For simplicity, only the foreground branch is illustrated in Table~\ref{tab:CSNet architectures} as the background one adopts the same architecture.  Note that this network simultaneously handles two complementary tasks as well as their relationships, which is denoted as Complementary Segmentation Network (CSNet). The parameters of the shared trunk are initialized from the ResNet50 networks~\cite{he2016deep}, which are used to extract low to high-level features that are shared by foreground objects and background distractors.
	We remove the pooling layer and the fully connected layer after \texttt{RELU} layer of \texttt{res5c}, and introduce two pooling blocks (see Fig.~\ref{Fig:CCNN}) to provide features from additional levels and reduce parameters. In order to integrate both local and global context, we sum up different levels of features output by layer \texttt{Res3}, \texttt{Res4} and \texttt{Res5} and two pooling blocks by appropriate up/down-sampling operations. After that, a residual block with a 3x3 \texttt{CONV} layer and a 1x1 \texttt{CONV} layer is used to post-process the integrated features as well as increase their nonlinearity. Finally, the shared trunk takes a $320\times{}320$ image as the input, and outputs a $40\times40$ feature map with 512 channels.
	%conduct dilated convolution in all subsequent \texttt{CONV} layers to extend the receptive fields without loss of resolution~\cite{yu2016dilatedconv}. the last two fully connected layers are converted into convolutional layers with $7\times7$ and $1\times{}1$ kernels, respectively.  , which are further compressed into a feature map with 128 channels by using a \texttt{CONV} layer with $1\times1$ kernels

	\begin{table}[!htp]
		\small
		\centering{
			\caption{Main configurations for CSNet. Note that x and y are integers in the range [1, 5].}
			\label{tab:CSNet architectures}
			\begin{tabular}{l|l|l}
				\toprule\rule{0pt}{8pt}
				%			\multirow{2}{c}{type name}  & \multirow{2}{c}{patch size/stride/pad \\ /dilation/group} &  \multirow{2}{c}{input size} & \multirow{2}{c}{output size} \tabularnewline
				\multirow{2}{*}{type name}  & {patch size/stride/} &   \multirow{2}{*}{output size} \tabularnewline
				~~&pad/dilation/group&~~ \\
				\midrule\rule{0pt}{8pt}
				
				Conv1\_pb1 & 3x3/1/1/1/32 &  10x10x256 \\
				\hline\rule{0pt}{10pt}		
				{Conv2\_pb1} & 1x1/1/0/1/1 &  10x10x2048 \\
				\hline\rule{0pt}{10pt}
				Pool1 & avg pool, 2x2, stride 2 &  5x5x2048 \\
				\hline \rule{0pt}{10pt}
				Conv3\_pb1  & 3x3/1/1/1/32  & 5x5x256 \\
				\hline\rule{0pt}{10pt}
				Conv4\_pb1  & 1x1/1/0/1/1  & 5x5x2048 \\
				\hline\rule{0pt}{10pt}
				Conv1\_pb2 & 3x3/1/1/1/32 &  5x5x256 \\
				\hline\rule{0pt}{10pt}		
				{Conv2\_pb2 } & 1x1/1/0/1/1 &  5x5x2048 \\
				\hline\rule{0pt}{10pt}
				Pool2 & avg pool, 2x2, stride 2 &  3x3x2048 \\
				\hline \rule{0pt}{10pt}
				Conv3\_pb2 & 3x3/1/1/1/32  & 3x3x256 \\
				\hline\rule{0pt}{10pt}
				Conv4\_pb2 & 1x1/1/0/1/1  & 3x3x2048 \\
				\hline\rule{0pt}{10pt}
				
				Interp1  & bilinear upsampling  & 40x40x1024 \\
				\hline\rule{0pt}{10pt}
				Conv1 & 1x1/1/0/1/1  &  40x40x512 \\
				\hline\rule{0pt}{10pt}
				Interp2 & bilinear upsampling  & 40x40x2048 \\
				\hline\rule{0pt}{10pt}
				Conv2 & 1x1/1/0/1/1  &  40x40x512 \\
				\hline\rule{0pt}{10pt}
				Interp3 & bilinear upsampling  & 40x40x2048 \\
				\hline\rule{0pt}{10pt}
				Conv3 & 1x1/1/0/1/1  &  40x40x512 \\
				\hline\rule{0pt}{10pt}
				Interp4 & bilinear upsampling  & 40x40x2048 \\
				\hline\rule{0pt}{10pt}
				Conv4 & 1x1/1/0/1/1  &  40x40x512 \\
				\hline\rule{0pt}{10pt}
				Conv5 & 3x3/1/1/1/32  &  40x40x256 \\
				\hline\rule{0pt}{10pt}
				Conv6 & 1x1/1/0/1/1  &  40x40x512 \\
				\hline\rule{0pt}{10pt}
				Conv7\_xf & 1x1/1/0/1/1  &  40x40x512 \\
				\hline\rule{0pt}{10pt}
				Conv8\_yf & 3x3/1/y/y/32  &  40x40x256 \\
				\hline\rule{0pt}{10pt}			
				Conv9f & 1x1/1/0/1/1  &  40x40x256 \\
				\hline\rule{0pt}{10pt}
				Conv10f & 3x3/1/1/1/8  &   40x40x64 \\
				\hline\rule{0pt}{10pt}
				Deconv1f & 3x3/4/1/1/1  &  161x161x1 \\
				\bottomrule
			\end{tabular}
		}
	\end{table}

	After the shared trunk, the features are fed into two separate branches that address two complementary tasks, \ie, foreground and background estimation. Note that the two branches share with the input, the architecture, but produce complementary outputs. In each branch, the shared features pass through a sequential of convolution blocks. These blocks all consist of $1 \times 1$ and $3 \times 3$ \texttt{CONV}s, but with different dilations. As such, we concatenate the output of each block to constitute feature maps at $40 \times 40$ resolution with $1280$ channels. These features, which have a wide range of spatial context and abstraction levels, are finally fed into several \texttt{CONV} layers for dimensional reduction and post-processing, and upsampled to produce output segmentation maps at size $161 \times 161$. With such designs, the foreground branch mainly focuses on detecting salient objects, while the background one suppresses distractors. In addition to the empirical loss defined in \eqref{eq:lossCCNN}, two additional losses~\eqref{eq:intersactionLoss}, \eqref{eq:unionLoss} are also adopted to penalize the conflicts and complementary degree of the output maps for more accurate predictions.
	
	%Alex认为group conv的方式能够增加 filter之间的对角相关性，而且能够减少训练参数，不容易过拟合，这类似于正则的效果

	In the training stage, we collect massive images with labeled salient objects from four datasets for image-based salient object detection~\cite{li2015mdf,MSRA10K,yan2013hierarchical,YangZLRY13Manifold}. We down-sample all images to $320\times320$ and their ground-truth saliency maps into $161\times161$. For the pretrained ResNet50 trunk the learning rate is set to $5\times10^{-7}$, while for the two branches they are $5\times10^{-6}$. We train the network with a mini-batch of $4$ images, using SGD optimizer with momentum $0.9$ and weight decay $0.0005$.

	\section{Efficient Temporal Propagation with Neighborhood Reversible Flow}\label{sec:NRF}
	
	The per-frame initialization of foregroundness and backgroundness can only provide a location prediction of the primary objects and background distractors at the spatial domain. However, the concept of primary objects is defined from a more global spatiotemporal perspective, not only salient in intra-frame but also consistent in inter-frame and throughout the whole video. Just as mentioned earlier, the primary video object should be spatiotemporally consistent, \ie, the saliency foreground regions should not change dramatically along the time dimension.
	This implies that there still exists a large gap between the frame-based initialization results and the video-based primary objects. Therefore, we need to further infer the primary objects that consistently pop-out in the whole video~\cite{VOSDataset} according to the spatiotemporal correspondence of visual signals. In this process, two key challenges need to be addressed, including:
	
	1)~how to find the most reliable correspondences between various (nearby or far-away) frames?
	
	2)~how to infer out the consistent primary objects based on spatiotemporal correspondences and the initialization results?
	
	To address these two challenges, we propose a neighborhood reversible flow algorithm to find and propagate neighborhood reversible subset from inter-frames. Details of our solutions will be discussed in the following part of this section.
	%relevancy

	\subsection{Neighborhood Reversible Flow}\label{sec:NRF-1}
	The proposed Neighborhood Reversible Flow (NRF) propagates information along reliable correspondences established among several key frames of the video, thus preventing errors to be accumulated fast and involving larger temporal windows for more effective context exploitation. Instead of pixel-level correspondence, NRF operates on superpixels to achieve region-level matching and higher computational efficiency.
	
	%Typically, the highly reliable optical flow can be constructed between adjacent frames, while their highly redundant visual content may prevent subsequent operations such as temporal foregroundness propagation. On the contrary, far-away frames can often bring more valuable cues on how to pop-out primary objects and suppress distractors. Unfortunately, the pixel-wise optical flow may suffer from large displacement in handling far-away frames, resulting in unreliable temporal correspondence.
	
	%In particular, finding the correspondences between the visual subsets of all frames can be very time-consuming.
	
	%To address the problems,
	%which is inspired by the concept of neighborhood reversibility in image search~\cite{jegou2010accurate}.
	Given a video $\mb{V}=\{\mc{I}_u\}_{u=1}^K$, we first apply the SLIC algorithm~\cite{achanta2012slic} to divide a frame $\mc{I}_u$ into $N_u$ superpixels, denoted as $\{\mc{O}_{ui}\}$. For each superpixel, we compute its average RGB, Lab and HSV colors as well as the horizontal and vertical positions. These features are then normalized into the same dynamic range $[0,1]$.
	
	Based on the features, we need to address two fundamental problems: 1) how to measure the correspondence between a superpixel $\mc{O}_{ui}$ from the frame $\mc{I}_u$ and a superpixel $\mc{O}_{vj}$ from the frame $\mc{I}_v$, and 2) which frames should be referred for a given frame? ~Inspired by the concept of neighborhood reversibility in image search~\cite{jegou2010accurate}, we can compute the pair-wise $\ell1$ distances between $\{\mc{O}_{ui}\}_{i=1}^{N_u}$ and $\{\mc{O}_{vj}\}_{j=1}^{N_v}$. After that, we denote the $k$ nearest neighbors of $\mc{O}_{ui}$ in the frame $\mc{I}_{v}$ as $\mc{N}_k(\mc{O}_{ui}|\mc{I}_v)$. As a consequence, two superpixels $\mc{O}_{ui}$ and $\mc{O}_{vj}$ are $k$-neighborhood reversible if they reside in the list of $k$ nearest neighbors of each other. That is,
	\begin{equation}\label{eq:reversable}
	\mc{O}_{ui}\in\mc{N}_k(\mc{O}_{vj}|\mc{I}_u) ~ ~\text{and}~ ~\mc{O}_{vj}\in\mc{N}_k(\mc{O}_{ui}|\mc{I}_v).
	\end{equation}
	From \eqref{eq:reversable}, we find that the smaller $k$, the more tightly two superpixels are temporally correlated. Therefore, the correspondence between $\mc{O}_{ui}$ and $\mc{O}_{vj}$ can be measured as
	\begin{equation}\label{eq:temCorr}
	s_{ui,vj}=\left\{
	\begin{aligned}
	\exp(-&2k/k_0),  ~ ~& \text{if}~k\leq{}k_0 \\
	&0,                  ~ ~& \text{otherwise}
	\end{aligned}
	\right.
	\end{equation}
	where $k_0$ is a constant to suppress weak flow and $k$ is a variable. A small $k_0$ will build sparse correspondences between $\mc{I}_u$ and $\mc{I}_v$ (\eg, $k_0=1$), while a large $k_0$ will cause dense correspondences. In this study, we empirically set $k_0=15$ and represent the flow between $\mc{I}_u$ and $\mc{I}_v$ with a matrix $\bl{F}_{uv}\in\mb{R}^{N_u\times{}N_v}$, in which the component at $(i,j)$ equals to $f_{ui,vj}$. Note that we further normalize $\bl{F}_{uv}$ so that each row sums up to 1. Considering the highly redundant visual content between adjacent frames, for each video frame $\mc{I}_u$ we pick up its adjacent keyframes $\{\mc{I}_t|t\in\mb{T}_u\}$ to ensure sufficient variation in content and depict reliable temporal correspondences. In this paper, we refer the interval $d_k$ of annotated video frames, which usually contain most critical information of the whole video, to determine the interval of adjacent keyframes. Later, we estimate the flow matrixes between a frame $\mc{I}_u$ and the frames $\{\mc{I}_t|t\in\mb{T}_u\}$, where $\mb{T}_u$ can be empirically set to $\{u-2\times d_k,u-d_k,u+d_k,u+2\times d_k\}$.
	
	\subsection{Temporal Propagation of Spacial Features}\label{sec:NRF-2}

	The flow $\{\bl{F}_{uv}\}$ depicts how superpixels in various frames are temporally correlated, which can be used to further propagate the spatial foregroundness and backgroundness. Typically, such temporal refinement can obtain impressive performance by solving a complex optimization problem with constraints like spatial compactness and temporal consistency. However, the time cost will also grow surprisingly high~\cite{zhang2015sosd}. Considering the requirement of efficiency in many real-world applications, we propose to minimize an objective function that has analytic solution. For a superpixel $\mc{O}_{ui}$, its foregroundness $x_{ui}$ and backgroundness $y_{ui}$ can be initialized as
	\begin{equation}
	x_{ui}=\frac{\sum_{p\in\mc{O}_{ui}}\bl{X}_u(p)}{|\mc{O}_{ui}|},~y_{ui}=\frac{\sum_{p\in\mc{O}_{ui}}\bl{Y}_u(p)}{|\mc{O}_{ui}|},
	\end{equation}
	where $p$ is a pixel with foregroundness $\bl{X}_u(p)$ and backgroundness $\bl{Y}_u(p)$. $|\mc{O}_{ui}|$ is the area of $\mc{O}_{ui}$. For the sake of simplification, we represent the foregroundness and backgroundness scores of all superpixels in the $u$th frame with column vectors $\bl{x}_u$ and $\bl{y}_u$, respectively. As a result, we can propagate such scores from $\mc{I}_v$ to $\mc{I}_u$ according to $\bl{F}_{uv}$:
	\begin{equation}\label{eq:foreProp}
	\bl{x}_{u|v}=\bl{F}_{uv}\bl{x}_{v},~\bl{y}_{u|v}=\bl{F}_{uv}\bl{y}_{v}, ~~\forall v\in{}\mb{T}_u.
	\end{equation}
	%We can see that strong flows tend to enforce the inter-frame consistency of primary objects.
	After the propagation, the foregroundness vector $\hat{\bl{x}}_{u}$ and backgroundness vector $\hat{\bl{y}}_{u}$ can be refined by solving
	\begin{equation}\label{eq:optObj}
	\begin{split}
	\hat{\bl{x}}_{u}&=\arg\min_{\bl{x}}~\|\bl{x}-\bl{x}_{u}\|_2^2+\lambda_c{}\sum_{v\in{}\mb{T}_u}\|\bl{x}-\bl{x}_{u|v}\|_2^2,\\
	\hat{\bl{y}}_{u}&=\arg\min_{\bl{y}}~\|\bl{y}-\bl{y}_{u}\|_2^2+\lambda_c{}\sum_{v\in{}\mb{T}_u}\|\bl{y}-\bl{y}_{u|v}\|_2^2,
	\end{split}
	\end{equation}
	where  $\lambda_c$ is a positive constant whose value is empirically set to 0.5. Note that we adopt only the $\ell2$ norm in \eqref{eq:optObj} so as to efficiently compute an analytic solution
	\begin{equation}\label{eq:analySolution}
	\begin{split}
	\hat{\bl{x}}_{u}= \frac{1}{1+\lambda_c \cdot |\mb{T}_u|}\left(\bl{x}_{u}+\lambda_c{}\sum_{v\in{}\mb{T}_u}\bl{F}_{uv}\bl{x}_{v}\right),\\
	\hat{\bl{y}}_{u}= \frac{1}{1+\lambda_c \cdot |\mb{T}_u|}\left(\bl{y}_{u}+\lambda_c{}\sum_{v\in{}\mb{T}_u}\bl{F}_{uv}\bl{y}_{v}\right).
	\end{split}
	\end{equation}
	By observing \eqref{eq:foreProp} and \eqref{eq:analySolution}, we find that the propagation process is actually calculating the average foregroundness and backgroundness scores within a local temporal slice under the guidance of neighborhood reversible flow. After the temporal propagation, we turn superpixel-based scores into pixel-based ones as
	\begin{equation}\label{fig:seg2pix}
	\bl{M}_u(p)=\sum_{i=1}^{N_u}\delta(p\in{}\mc{O}_{ui})\cdot{}\hat{x}_{ui}\cdot(1-\hat{y}_{ui}),
	\end{equation}
	where $\bl{M}_u$ is the importance map of $\mc{I}_u$ that depict the presence of primary objects. $\delta(p\in{}\mc{O}_{ui})$ is an indicator function which equals to 1 if $p\in{}\mc{O}_{ui}$ and 0 otherwise. Finally, we calculate an adaptive threshold which equals to the 20\% of the maximal pixel importance to binarize each frame, and a morphological closing operation is then performed to fill in the black area in the segmented objects.

	\begin{table*}[!t]
		\small
		\centering{
			\caption{Performances of our approach and 18 models. Bold and underline indicate the 1st and 2nd performance in each column. ImageN: Image-based \& Non-deep. ImageD: Image-based \& Deep. }
			\label{tab:performances}
			\begin{tabular}{c@{}r|cccc|cccc|cccc}
				\toprule
				\multicolumn{2}{c|}{\multirow{2}*{Models}}  & \multicolumn{4}{c|}{\bl{SegTrackV2} (14 videos)} &  \multicolumn{4}{c|}{\bl{Youtube-Objects} (127 videos)} & \multicolumn{4}{c}{\bl{VOS} (200 videos)}  \tabularnewline
				&                                 &~mAP~     &~mAR~     &~$F_\beta$&~mIoU~     &~mAP~     & ~mAR~    &~$F_\beta$&~mIoU~     &~mAP~     & ~mAR~    &~$F_\beta$&~mIoU              \tabularnewline
				\midrule
				\bl{\multirow{7}{*}{\rotatebox{90}{ImageN}}}
				&~~\bl{DRFI}~\cite{JiangWYWZL13}	 &    .464 &    .775 &    .511 &    .395 &    .542 &    .774 &    .582 &    .453 &    .597 &    .854 &    .641 &    .526 \tabularnewline
				&~~\bl{RBD}~\cite{zhu2014saliency}	 &    .380 &    .709 &    .426 &    .305 &    .519 &    .707 &    .553 &    .403 &    .652 &    .779 &    .677 &    .532 \tabularnewline
				&~~\bl{BL}~\cite{Tong2015BL} 	     &    .202 &\bl{.934}&    .246 &    .190 &    .218 &\bl{.910}&    .264 &    .205 &    .483 &\bl{.913}&    .541 &    .450 \tabularnewline
				&~~\bl{BSCA}~\cite{Qin2015BSCA}	 &    .233 &    .874 &    .280 &    .223 &    .397 &    .807 &    .450 &    .332 &    .544 &    .853 &    .594 &    .475 \tabularnewline
				&~~\bl{MB+}~\cite{Zhang2015MB}	     &    .330 &\ul{.883}&    .385 &    .298 &    .480 &    .813 &    .530 &    .399 &    .640 &    .825 &    .675 &    .532 \tabularnewline
				&~~\bl{MST}~\cite{tu2016mst}	     &    .450 &    .678 &    .488 &    .308 &    .538 &    .698 &    .568 &    .396 &    .658 &    .739 &    .675 &    .497 \tabularnewline
				&~~\bl{SMD}~\cite{peng2016smd}	     &    .442 &    .794 &    .493 &    .322 &    .560 &    .730 &    .592 &    .424 &    .668 &    .771 &    .690 &    .533 \tabularnewline
				%&~~                                  &         &         &         &         &         &         &         &         &         &         &         &         \tabularnewline
				\hline
				\bl{\multirow{6}{*}{\rotatebox{90}{ImageD}}}
				&~~\bl{MDF}~\cite{li2015mdf}	     &    .573 &    .634 &    .586 &    .407 &    .647 &    .776 &    .672 &    .534 &    .601 &    .842 &    .644 &    .542 \tabularnewline
				&~~\bl{ELD}~\cite{lee2016eld}	     &    .595 &    .767 &    .627 &    .494 &    .637 &    .789 &    .667 &    .531 &    .682 &    .870 &    .718 &    .613 \tabularnewline
				&~~\bl{DCL}~\cite{li2016dcl}	     &    .757 &    .690 &    .740 &    .568 &    .727 &    .764 &    .735 &    .587 &\ul{.773}&    .727 &\ul{.762}&    .578 \tabularnewline
				&~~\bl{LEGS}~\cite{wang2015legs}	 &    .420 &    .778 &    .470 &    .351 &    .549 &    .776 &    .589 &    .450 &    .606 &    .816 &    .644 &    .523 \tabularnewline
				&~~\bl{MCDL}~\cite{zhao2015mcdl}	 &    .587 &    .575 &    .584 &    .424 &    .647 &    .613 &    .638 &    .471 &    .711 &    .718 &    .713 &    .581 \tabularnewline
				&~~\bl{RFCN}~\cite{wang2016rfcn}	 &    .759 &    .719 &    .749 &    .591 &\ul{.742}&    .750 &\ul{.744}&\ul{.592}&    .749 &    .796 &    .760 &\ul{.625}\tabularnewline
				%&~~                                  &         &         &         &         &         &         &         &         &         &         &         &         \tabularnewline
				\hline
				\bl{\multirow{5}{*}{\rotatebox{90}{Video-based}}}
				&~~\bl{NLC}~\cite{Faktor2014nlc}	 &\bl{.933}&    .753 &\bl{.884}&\bl{.704}&    .692 &    .444 &    .613 &    .369 &    .518 &    .505 &    .515 &    .364 \tabularnewline
				&~~\bl{ACO}~\cite{jang2016primary}	 &\ul{.827}&    .619 &    .767 &    .551 &    .683 &    .481 &    .623 &    .391 &    .706 &    .563 &    .667 &    .478 \tabularnewline
				&~~\bl{FST}~\cite{papazoglou2013fst}&    .792 &    .671 &    .761 &    .552 &    .687 &    .528 &    .643 &    .380 &    .697 &    .794 &    .718 &    .574 \tabularnewline
				&~~\bl{SAG}~\cite{wang2015sag}	     &    .431 &    .819 &    .484 &    .384 &    .486 &    .754 &    .529 &    .397 &    .538 &    .824 &    .585 &    .467 \tabularnewline
				&~~\bl{GF}~\cite{wang2015GF}	     &    .444 &    .737 &    .489 &    .354 &    .529 &    .722 &    .563 &    .407 &    .523 &    .819 &    .570 &    .436 \tabularnewline
				%&~~                                  &         &         &         &         &         &         &         &         &         &         &         &         \tabularnewline
				%\hline
				%&~\bl{OUR_pre}                           &    .809 &    .802 &\ul{.807}&\ul{.670}&\bl{.745}&    .798 &\bl{.756}&\bl{.633}&\bl{.789}&    \ul{.870} &\bl{.806}&\bl{.710} \tabularnewline
				\hline
				&~\bl{CSP}                           &    .789 &    .778 &\ul{.787}&\ul{.669}&\bl{.778}&\ul{.820}&\bl{.787}&\bl{.675}&\bl{.805}&    \ul{.910} &\bl{.827}&\bl{.747} \tabularnewline
				\bottomrule
			\end{tabular}
		}
	\end{table*}

	\section{Experiments}\label{sec:Experiments}
	In this section, we first illustrate experimental settings about datasets and evaluation metrics in Section 5.1.
	Then based on the datasets and metrics, we compare quantitatively our primary video object segmentation method with 18 state-of-the-art approaches in Section 5.2.
	After that, in Section 5.3 we further demonstrate the effectiveness of our approach by offering more detailed exploration and dissecting various parts of our approach as well as running time and failure cases.
	%\vspace{2cm}
	
	\subsection{Experimental Settings}\label{sec:Experimental Settings}
	
	We test the proposed approach on three widely used video datasets, while their ways in defining primary video objects are different. Details of these datasets are described as follows:
	
	\myPara{1)~SegTrack V2}~\cite{li2013segtrackv2} is a classic dataset in video object segmentation that are frequently used in many previous works. It consists of 14 densely annotated video clips with $1,066$ frames in total. Most primary objects in this dataset are defined as the ones with \emph{irregular motion patterns}.
	
	\myPara{2)~Youtube-Objects}~\cite{prest2012youtubeobjects} contains a large amount of Internet videos and we adopt its subset \cite{jain2014supervoxel} that contains 127 videos with $20,977$ frames. In these videos, $2,153$ keyframes are sparsely sampled and manually annotated with pixel-wise masks according to the video tags. In other words, primary objects in \bl{Youtube-Objects} are defined from the perspective of \emph{semantic attributes}.
	
	\myPara{3)~VOS}~\cite{VOSDataset} contains 200 videos with $116,093$ frames. On $7,467$ uniformly sampled keyframes, all objects are pre-segmented by 4 subjects, and the fixations of another 23 subjects are collected in eye-tracking tests. With these annotations, primary objects are automatically selected as the ones whose average fixation densities over the whole video fall above a predefined threshold. If no primary objects can be selected with the predefined threshold, objects that receive the highest average fixation density will be treated as the primary ones. Different from \bl{SegTrack V2} and \bl{Youtube-Objects}, primary objects in \bl{VOS} are defined from the perspective of \emph{human visual attention}.
	
	%&~~                                  &         &         &         &         &         &         &         &         &         &         &         &         \tabularnewline

	On these three datasets, the proposed approach, denoted as \bl{CSP}, is compared with 18 state-of-the-art models for primary and salient object segmentation, including:
	
	1)~Image-based \& Non-deep (7): \bl{RBD}~\cite{zhu2014saliency}, \bl{SMD}~\cite{peng2016smd}, \bl{MB+}~\cite{Zhang2015MB}, \bl{DRFI}~\cite{JiangWYWZL13}, \bl{BL}~\cite{Tong2015BL}, \bl{BSCA}~\cite{Qin2015BSCA}, \bl{MST}~\cite{tu2016mst}.
	
	2)~Image-based \& Deep (6): \bl{ELD}~\cite{lee2016eld}, \bl{MDF}~\cite{li2015mdf}, \bl{DCL}~\cite{li2016dcl}, \bl{LEGS}~\cite{wang2015legs}, \bl{MCDL}~\cite{zhao2015mcdl} and \bl{RFCN}~\cite{wang2016rfcn}.
	
	3)~Video-based (5): \bl{ACO}~\cite{jang2016primary}, \bl{NLC}~\cite{Faktor2014nlc}, \bl{FST}~\cite{papazoglou2013fst}, \bl{SAG}~\cite{wang2015sag} and \bl{GF}~\cite{wang2015GF}.
	
	In the comparisons, we adopt two sets of evaluation metrics, including the Intersection-over-Union (IoU) and the Precision-Recall-$F_\beta$. Similar to \cite{VOSDataset}, the precision, recall and IoU scores are first computed on each video and finally averaged over the whole dataset so as to generate the mean Average Precision (mAP), mean Average Recall (mAR) and mean Average IoU (mIoU). In this manner, the influence of short and long videos can be balanced. Furthermore, a unique $F_\beta$ score can be obtained based on mAR, mAP and a parameter $\beta$, the square of which is set as 0.3 to emphasize precision more than recall in the evaluation.
	
	%%In the comparisons, we adopt two sets of evaluation metrics, including the Intersection-over-Union (IoU) and the Precision-Recall-$F_\beta$. Similar to \cite{VOSDataset}, the precision, recall and IoU scores are first computed on each frame, which are then separately averaged over each video and finally averaged over the whole dataset so as to generate the mean Average Precision (mAP), mean Average Recall (mAR) and mean Average IoU (mIoU). In this manner, the influence of short and long videos can be balanced. Furthermore, a unique $F_\beta$ score is computed based on mAR and mAP:
	
	%\begin{equation}
	%F_\beta=\frac{(1+\beta^2)\cdot\text{mAP}\cdot{}\text{mAR}}{\beta^2\cdot\text{mAP}+\text{mAR}},
	%\end{equation}
	%where we set $\beta^2=0.3$ as in many previous works to emphasize precision more than recall in the evaluation. Among these metrics, mAR and mAP explain why a model performs impressive or fail from the complementary perspectives of recall and precision, which can help to provide more insights beyond a single mIoU score.
	
	\subsection{Comparisons with State-of-the-art Models}\label{sec:Comparisons}
	
	\begin{figure*}[!t]
		\centering {
			\includegraphics[width=1.0\textwidth]{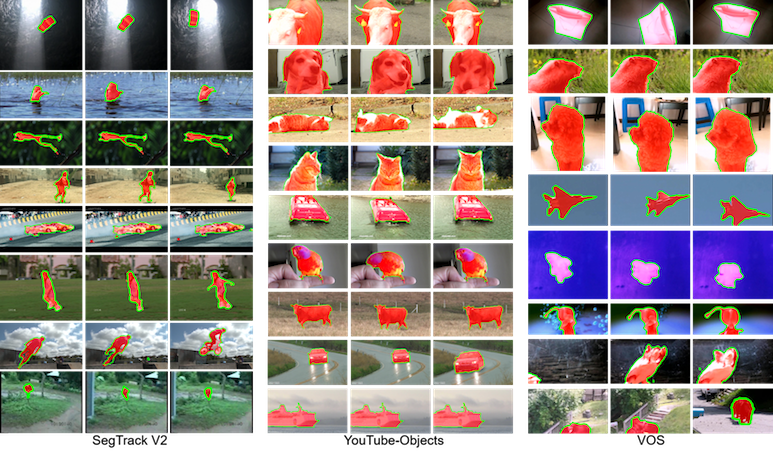}}
		\caption{Representative results of CSP. Red masks are the ground-truth and green contours are the segmention objects.}
		\label{Fig:examples}
	\end{figure*}
	
	%\vspace{5cm}
	
	The performances of our approach and 18 state-of-the-art models on three video datasets are shown in Table~\ref{tab:performances}. Some representative results of our approach are demonstrated in Fig.~\ref{Fig:examples}.
	From Table~\ref{tab:performances}, we find that on \bl{Youtube-Objects} and \bl{VOS} such larger datasets our approach obtains the best $F_\beta$ and mIoU scores, while on \bl{SegTrack V2} our approach ranks the second places (worse than \bl{NLC}). This can be explained by the fact that \bl{SegTrack V2} contains only 14 videos, among which most primary objects have irregular motion patterns. Such videos often perfectly meet the assumption of \bl{NLC} on motion patterns of primary objects, making it the best approach on \bl{SegTrack V2}. However, when the scenarios being processed extend to datasets like \bl{VOS} that are constructed without such ``constraints'' on motion patterns, the performance of \bl{NLC} drops sharply as its assumption may sometimes fail (\eg, \bl{VOS} contains many videos only with static primary objects and distractors as well as slow camera motion, see Fig.~\ref{Fig:examples}). These results further validate that it is quite necessary to conduct comparisons on larger datasets with daily videos (like \bl{VOS}) so that models with various kinds of assumptions can be fairly evaluated.
	
	Moreover, there exist some approaches (\bl{BL} and \bl{MB+}) on the three datasets that outperform our approach in recall, and some other approaches (\bl{NLC}, \bl{ACO} and \bl{FST}) may achieve better or comparable precision with our approach on \bl{SegTrack V2}. However, the other evaluation scores of the approaches are much worse than our method on the three datasets. That is, none of these approaches simultaneously outperforms our approach in both recall and precision so that our approach often have better overall performance, especially on larger datasets. This may imply that the proposed approach is more balanced than previous works. By analyzing the results on the three datasets, we find that this phenomenon may be caused by the conduction of complementary tasks in CSNet. By propagating both foregroundness and backgroundness, some missing foreground information may be retrieved, while the mistakenly popped-out distractors can be suppressed again, leading to balanced recall and precision.

	%detail CCNN CCNN*
	\begin{table*}[!t]
		\small
		\centering{
			\caption{Detail performances of our approaches. The first test group is our previous work in~\cite{li2017primary} and the second group is our current work. V-Init/R-Init. : corresponding results initialized by previous/current network. FG (FGp)/BG (BGp): foreground/background estimation with (without) the constraint of complementary loss.  NRF (NRFp): Neighborhood Reversible Flow (with multi-test). CE: cross-entropy. Comple.: complementary loss. mT: mean temporal stability metric, the smaller the better. Bold and underline indicate the 1st and 2nd performance in each column.}
			\label{tab:ourperformances}
			\begin{tabular}{c@{}l|c|cc|c|c|ccccc}
				\toprule
				\multicolumn{2}{c|}{\multirow{2}*{Test cases}}  & \bl{Backbone}  &  \multicolumn{2}{c|}{\bl{Objective} } & ~~~  & ~~~  &\multicolumn{5}{c}{\bl{Evaluation} }  \tabularnewline
				&                                 &VGG16/ResNet50  &CE  &Comple.     & NRF   & multi-test   &~mAP~     & ~mAR~    &~$F_\beta$  &~mIoU   &~mT~           \tabularnewline
				\midrule
				\bl{\multirow{4}{*}{\rotatebox{90}{ }}}
				&~~\bl{V-Init. FG}	     &  VGG16    &  \checkmark & ~~           &  ~~        & ~~&   .750 &    .879 &    .776 &    .684 &    .117  \tabularnewline
				&~~\bl{V-Init. BG}	     &  VGG16    &  \checkmark & ~~           &  ~~        & ~~&   .743 &    .884 &    .771 &    .680 &    .117   \tabularnewline
				&~~\bl{V-Init. (FG+BG)} &  VGG16    &  \checkmark & ~~           &  ~~        & ~~&   .791 &    .834 &    .800 &    .689 &    .121  \tabularnewline
				&~~\bl{V-Init. +NRF}	 &  VGG16    &  \checkmark & ~~           & \checkmark & ~~&   .789 &    .870 &    .806 &    .710 &    .109   \tabularnewline
				\hline
				\bl{\multirow{8}{*}{\rotatebox{90}{ }}}
				&~~\bl{R-Init. FG}	     &  ResNet50    &   \checkmark  &~          &  ~~        & ~~&   .763 &    .901 &    .791 &    .710 &    .128  \tabularnewline
				&~~\bl{R-Init. BG}	     &  ResNet50    &   \checkmark  &~          &  ~~        & ~~&   .764 &    .899 &    .791 &    .711 &    .128   \tabularnewline
				&~~\bl{R-Init. (FG+BG)}    &  ResNet50    &   \checkmark  &~          &  ~~        & ~~&  \ul{.808} &    .863 &    .820 &    .724 &    .127  \tabularnewline
				&~~\bl{R-Init. FGp}	     &  ResNet50    &  \checkmark   &\checkmark &  ~~        & ~~&   .768 &  \ul{.925} &    .800 &    .726 &    .124  \tabularnewline
				&~~\bl{R-Init. BGp}	     &  ResNet50    &  \checkmark   &\checkmark &  ~~        & ~~&   .763 &    \bl{.927} &    .796 &    .723 &    .124   \tabularnewline
				&~~\bl{R-Init. (FGp+BGp)}  &  ResNet50    &   \checkmark  &\checkmark &  ~~        & ~~&   \bl{.814} &    .883 &  \bl{.829} &    .739 &    .122  \tabularnewline
				&~~\bl{R-Init. +NRF}	     &  ResNet50    &   \checkmark  &\checkmark &\checkmark  & ~~&   .803 &    .901 &    .824 &  \ul{.739} &   \ul{.108}   \tabularnewline
				&~~\bl{R-Init. +NRFp}	     &  ResNet50    &   \checkmark  &\checkmark &\checkmark  & \checkmark&   .805 &    .910 &  \ul{.827} &    \bl{.747} &    \bl{.097}   \tabularnewline
				\bottomrule
			\end{tabular}
		}
	\end{table*}

	From Table~\ref{tab:performances}, we also find that there exist inherent correlations between salient image object detection and primary video object segmentation. As shown in Fig.~\ref{Fig:examples}, primary objects are often the most salient ones in many frames, which explains the reason that deep models like \bl{ELD}, \bl{RFCN} and \bl{DCL} outperforms many video-based models like \bl{NLC}, \bl{SAG} and \bl{GF}. However, there are several key differences between the two problems. First, primary objects may not always be the most salient ones in all frames (as shown in Fig.~\ref{Fig:primary}). Second, inter-frame correspondences provide additional cues for separating primary objects and distractors, which depict a new way to balance recall and precision. Third, primary objects may be sometimes close to video boundary due to camera and object motion, making the boundary prior widely used in many salient object detection models no valid any more (\eg, the bear in the last row of the last column of Fig.~\ref{Fig:examples}). Last but not least, salient object detection needs to distinguish a salient object from a fixed set of distractors, while primary object segmentation needs to consistently pop-out the same primary object from a varying set of distractors. To sum up, primary video object segmentation is a more challenging task that needs to be further explored from the spatiotemporal perspective.

	\subsection{Detailed Performance Analysis}\label{sec:Analysis}
	
	Beyond performance comparison, we also conduct several experiments on \bl{VOS}, the largest one of the three datasets, to find out how the proposed approach works in segmenting primary video objects. Moreover, an additional metric, \ie, temporal stability measure $T$~\cite{Perazzi2016}, is applied to evaluating the relevant aspect in primary video objects segmentation in addition to the aforementioned four metrics. After all, mIoU only measures how well the pixels of two masks match, while $F_\beta$ measures the accuracy of contours. None of them consider the temporal aspect. However, video objects segmentation is conducted in spatiotemporal dimensions. So the additional temporal stability measure is a appropriate choice to evaluate the temporal consistency of segmentation results. The main quantifiable results can be found in Table.~\ref{tab:ourperformances}. In Table.~\ref{tab:ourperformances}, the first group is our previous work in \cite{li2017primary} and the second group is our current work extended from \cite{li2017primary}. In order to illustrate the effect of each component in our approach, the two group of tests are based on the same parameters except for the last case R-Init.+NRFp, which is the final test result in generally by using some data argumentation and parameter adjustment on the base of case R-Init.+NRF.

	\subsubsection{Performance of Complementary CNNs}\label{sec:Analysis1 CCNN}
	
	%	The visual effects of foreground maps and background maps initialized by the proposed complementary CNN branches have been shown in Fig.~\ref{Fig:complementary}. To further verify the effectiveness, some detail analysis will be given in this section by evaluating main components such as the two complementary CNN branches and complementary loss \eqref{eq:priorCCNN}.
	
	In this section, some detail analysis will be given to further verify the effectiveness of the proposed complementary CNN branches and complementary loss.
	
	\bl{Impact of two complementary branches}.~~To explore the impact of two complementary network branches, we evaluate the foreground maps and background maps initialized by the two complementary branches, as well as their fusion maps. As shown in Table.~\ref{tab:ourperformances}, in the first group the evaluation scores of case V-Init. FG and V-Init. BG are equally matched for their same branch structure, while the ones of their fusion maps are increased at different degree, which suggests that the complementary characteristics of initialized foreground maps and background maps can contribute to and constrain each other to generate more accurate prediction.
	Then what will happen if we abandon the background branch? To this end, we conduct two additional experiments in our previous work~\cite{li2017primary}. First, if we cut down the background branch and retrain only the foreground branch, the final performance decreases by about 0.9\%. Second, if we re-train a network with two foreground branches, the final $F_\beta$ and mIoU scores decrease from 0.806 to 0.800 and 0.710 to 0.700, respectively. These experiments indicate that, beyond learning more weights, the background branch does have learned some useful cues that are ignored by the foreground branch, which are expected to be high-level visual patterns of typical background distractors. These results also validate the idea of training deep networks by simultaneously handling two complementary tasks.
	
	Therefore, network structure with two similar branches are still adopted in this extension work. What is different is that the new network structure is assisted with more designment based on the deeper and more effective ResNet50 instead of simple VGG16.
	%For the sake of differentiation, we express the new network structure as CCNN+ (\ie, the forementioned CSNet).
	From Table.~\ref{tab:ourperformances}, we can find that the initialized results are distinctly improved when the backbone VGG16 is replaced by ResNet50. The aforementioned four evaluation scores are all increased, \eg, the $F_\beta$ and mIoU scores increase from 0.776 to 0.791 and 0.684 to 0.710, respectively, although the temporal stability performance is affected. This reveals the better performance of our new network structure, and at the same time hints a fact that a favourable per-frame initialization cannot stand for a good video initialization because of the temporal consistency attribute in video. Thus, it is necessary to conduct optimization in temporal dimension, which will be explained in next subsection.
	
	\begin{figure}[!t]
		\centering {
			\includegraphics[width=1.0\columnwidth]{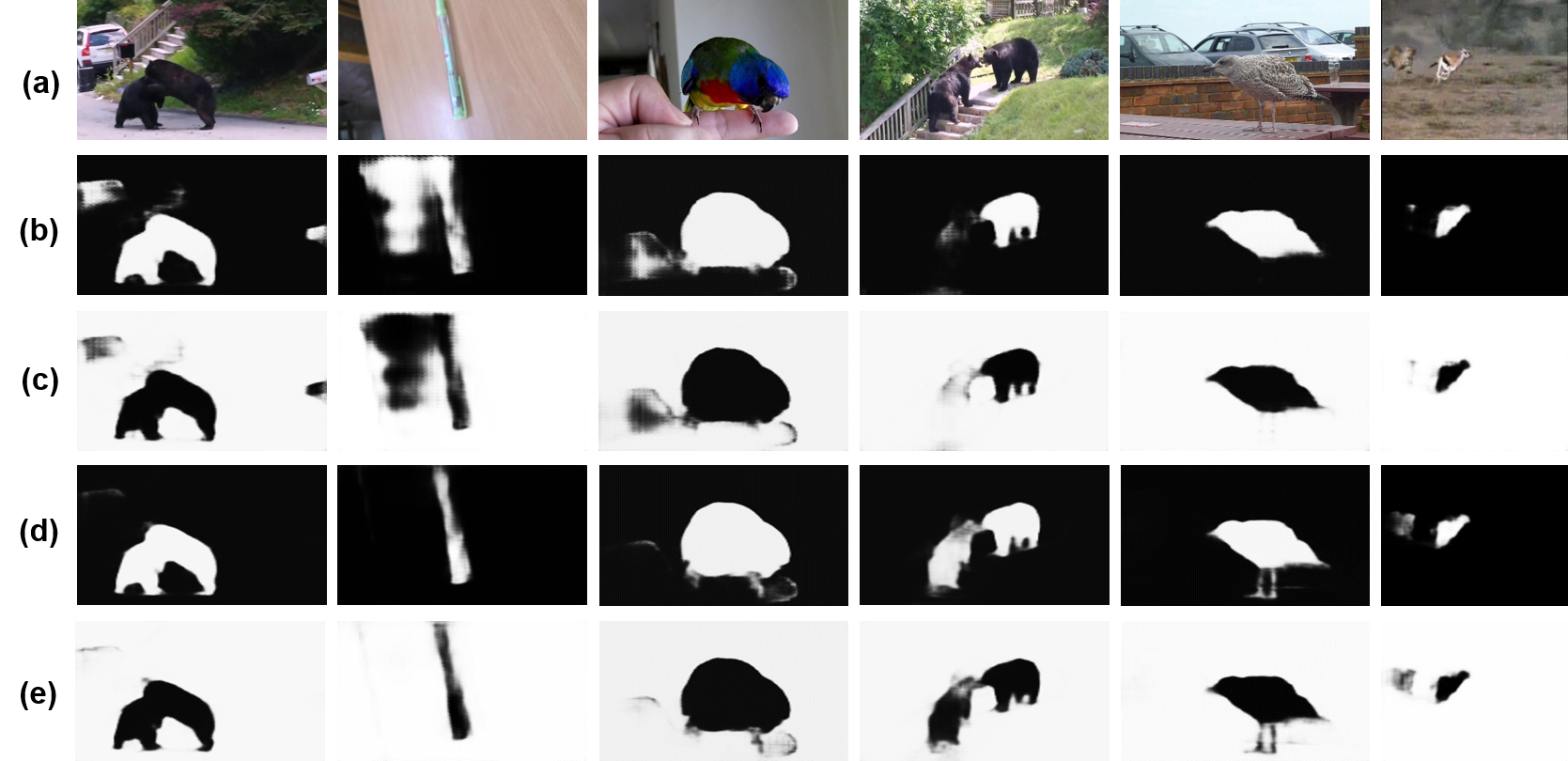}}
		\caption{Foreground and background maps initialized by CSNet as well as their interaction and union maps (a)~video frames, (b) foreground maps and (c) background maps generated by CSNet without the complementary loss. (d) foreground maps and (e) background maps generated by CSNet with the complementary loss. }
		\label{Fig:complementary}
	\end{figure}
	
	\bl{Effect of the complementary loss}.~~Except for the specific network, another main difference is that the two complementary CNN bracnches in CSNet are also constrained by our complementary loss. 	
	To verify the effectiveness, we optimize two sets of foreground and background prediction models based on the new network structure, one with the constraint of the penalty term in the objective function, and the other without. Based on the two sets of models, we can initialize a foreground and a background map for each video frame. The quantitative evaluations of initialization results respectively correspond to the cases R-Init. FGp/BGp/(FGp+BGp) and the cases R-Init. FG/BG/(FG+BG) in Table.~\ref{tab:ourperformances}.  From Table.~\ref{tab:ourperformances} we can find that compared to the predictions without penalty term constraint, the foreground and background models with the additional complementary loss can achieve better performance in predicting both foreground maps and background maps, shown as the better $F_\beta$ and mIoU scores. Moreover, some visual examples are shown in Fig.~\ref{Fig:complementary}. Obviously, if we only use the empirical loss~\eqref{eq:lossCCNN}, some background regions may be wrongly classified into foreground (\eg~the first three columns in Fig.~\ref{Fig:complementary}) while some foreground details may miss (\eg~the last three columns in Fig.~\ref{Fig:complementary}). By incorporating the additional complementary loss~\eqref{eq:priorCCNN}, these mistakes can be fixed (see Fig.~\ref{Fig:complementary}(d)(e)). Thus, the complementary loss is effective for boundary localizations and suppressing background distractors. These results validate the effectiveness of handling two complementary tasks with explicit consideration of their relationships.
	
	Combining the two differences, the $F_\beta$ and mIoU scores of initialization results output by our previous network CCNN (the case V-Init. (FG+BK)) increase by about 3.6\% and 7.2\%, \ie, from 0.800 to 0.829 and 0.689 to 0.739, respectively, with the increased mAP and mAR scores. This also means that the combination of complementary loss and two complementary branches of foreground and background are valid and ingenious.
	
	In particular, the complementary CNN branches in our two networks both show impressive performance in predicting primary video objects over the other 6 deep models when their pixel-wise predictions are directly evaluated on \bl{VOS}. By analyzing the results, we find that this may be caused by two reasons: 1)~using more training data, and 2)~simultaneously handling complementary tasks, whose effectiveness is just verified. To explore the first reason, we retrain CCNN on the same MSRA10K dataset used by most deep models. In this case, the $F_\beta$ (mIoU) scores of the foreground and background maps predicted by CCNN decrease to 0.747 (0.659) and 0.745 (0.658), respectively. Note that both branches still outperform \bl{RFCN} on \bl{VOS} in terms of mIoU (but $F_\beta$ is slightly worse).

	%------------
	%We conduct several experiments to assess the proposed complementary cost function. As shown in Table \ref{tab:iulossperformances}, both CCNN branches show impressive performance in predicting primary video objects when their pixel-wise predictions are directly evaluated as the other 6 deep models. Although the foreground/backgroundness conversion from pixel to superpixel may slightly decrease the precision and increase the recall, the overall precision increases by a large extent after the temporal propagation, at the cost of small decrease in recall. Considering that high precision is much more difficult to reach than high recall, such trade-off can bring increasing $F_\beta$ and mIoU scores.
	%
	%In particular, the performances of both the foregroundness and backgroundness branches outperform all the other 6 deep image-based salient object detection models on \bl{VOS}. By analyzing the results, we find that this may be caused by two reasons: 1)~using more training data, and 2)~simultaneously handling complementary tasks. To further explore the reasons, we retrain CCNN on the same MSRA10K dataset used by most deep models. In this case, the $F_\beta$ (mIoU) scores of the foregroundness and backgroundness maps predicted by CCNN will decrease to 0.747 (0.659) and 0.745 (0.658), respectively. Note that both branches still outperform \bl{RFCN} on \bl{VOS} in terms of mIoU (but $F_\beta$ is slightly worse).

	\subsubsection{Effectiveness of Neighborhood Reversible Flow}\label{sec:Analysis1 NRF}
	
	\begin{figure}[!t]
		\centering {
			\includegraphics[width=1.0\columnwidth]{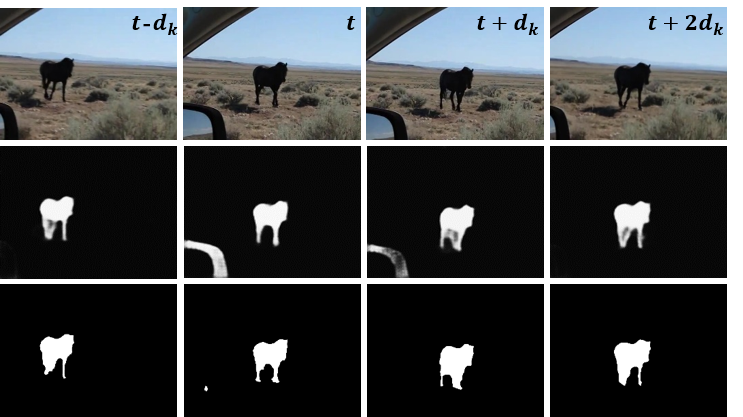}} %
		\caption{Performance of the proposed neighborhood reversible flow. The first row is the video squences from \bl{Youtube-Objects}~\cite{prest2012youtubeobjects}, the second row is corresponding initialized foreground maps, and the third row is optimized results by neighborhood reversible flow.}
		\label{Fig:nrfperform} %?????
	\end{figure}

	Through above complementary network branches, the salient foreground and background maps in intra-frame are well obtained, while the initialization operation cannot ensure temporal consistency of segmented objects, \eg, the initialized predictions by CSNet outperform the ones by CCNN in term of mAP, mAR, $F_\beta$ and mIoU, while become inferior in temporal stability mT.  Thus a thought of improving temporal relationship is proposed in section~\ref{sec:NRF}, \ie, finding and propagating the reliable inter-frame correspondences by applying the neighborhood reversible flow to make the consistent salient subsets enhanced and accidental distractors suppressed. Consequently, the final primary video objects with spatiotemporal consistency are yielded.
	
	\bl{Effectiveness of Neighborhood Reversible Flow}.~~To prove this thought, we compare the initialized results by CCNN (CSNet) with the optimized results (V-Init.(R-Init.)+NRF) by neighborhood reversible flow. As shown in Table~\ref{tab:ourperformances}, the temporal stability measure mT of optimization results in CCNN (CSNet) cases decrease from 0.121 (0.122) to 0.109 (0.108) comparing with the initialized predictions (V-Init. (FG+BK), R-Init. (FGp+BKp)). At the same time, the other evaluation scores are also improved, \eg, the mIoU score from 0.689 to 0.710. The superiority will become more obviously if we directly compare V-Init.+NRF with V-Init. FG,~\ie, conduct the fusion operation on foreground and background in the process of neighborhood reversible flow just like we really do. This means that by propagating the neighborhood reversible flow, the spatial subsets of primary objects in intra-frame can be refined from a temporal perspective and the inter-frame temporal consistency can be enhanced. Finally, the primary video objects with favourable spatiotemporal consistency can pop-out. As shown in Fig.~\ref{Fig:nrfperform}, the primary objects in most video frames are initialized as the horce, while the objects that only lasts for a short while are mistakenly classified into foreground due to their spatial saliency in certain frames. Fortunately, the distractors are well suppressed by the optimization of neighborhood reversible flow (see the third rows in Fig.~\ref{Fig:nrfperform}). Thus, via propagating salient cues in inter-frames, background objects could be effectively suppressed, only preserving the real primary one.

	To further demonstrate the effectiveness of neighborhood reversible flow, we test our approach with two new settings based on the CCNN. In the first setting, we replace the correspondence from Eq.~\eqref{eq:temCorr} to the Cosine similarity between superpixels. In this case, the $F_\beta$ and mIoU scores of our approach on \bl{VOS} drop to 0.795 and 0.696, respectively. Such performance is still better than the initialized foreground maps but worse than the performance when using the neighborhood reversible flow ($F_\beta$=0.806, mIoU = 0.710). This result indicates the effectiveness of neighborhood reversibility in temporal propagation.
	
	In the second setting, we set $\lambda_c=+\infty$ in Eq.~\eqref{eq:optObj}, implying that primary objects in a frame are solely determined by the foreground and background propagated from other frames. When the spatial predictions of each frame are actually ignored in the optimization process, the $F_\beta$ (mIoU) scores of our approach on \bl{VOS} only decrease from 0.806 (0.710) to 0.790 (0.693), respectively. This result proves that the inter-frame correspondences encoded in the neighborhood reversible flow are quite reliable for efficient and accurate propagation along the temporal dimension.
	\begin{table}[!t]
		\centering{
			\caption{Performance of superpixel wise initialization by CSNet on \bl{VOS}. FGp: foreground branch, BKp: background branch. Sup. is short for superpixel.}
			\label{tab:NRFsuperp}
			\begin{tabular}{c|ccccc}
				\toprule
				Step                          & mAP  & mAR  & $F_\beta$ & mIoU   & $T$  \tabularnewline
				\midrule
				%CCNN FG (Superpixel)               & .730(743) & .895(891) & .762(773) & .674(685)  & .134  \tabularnewline
				%CCNN BK (Superpixel)               & .722(736) & .901(895) & .756(767) & .669(681)  & .134  \tabularnewline
				%CCNN FG+BK (Superpixel)               & .795 & .845 & .806 & .701  & .131  \tabularnewline
				R-Init. FGp (Sup.)                   & .765 & .924 & .797 & .723  & .133  \tabularnewline
				R-Init BKp (Sup.)                   & .759 & .926 & .792 & .719  & .133  \tabularnewline
				R-Init FGp+BKp (Sup.)                   & .814 & .881 & .829 & .738  & .129  \tabularnewline
				\bottomrule
			\end{tabular}
		}
	\end{table}

	It's worth mentioning that in the previous initialization process, the predictions are all pixel-wise, while the temporal optimization via neighborhood reversible flow are conducted on superpixel wise foreground/background maps in order to reduce time consumption, \ie, the predictions need to be converted from the pixel to superpixel and finally converted to pixel. However, the superpixel wise predictions are relatively coarse and may affect the following process. To explore the effect, we convert the foreground/background maps and their fusion maps in cases R-Init. FGp/BKp from pixel-wise to superpixel wise, as shown in Table~\ref{tab:NRFsuperp}. Fortunately, both $F_\beta$ and mIoU scores of foreground (background) maps only slightly decrease by 0.003 (0.004) and the mT scores increase by 0.009, while the negative effect on fusion maps mainly manifests in mT scores, which can be improved by the propagation of neighborhood reversible flow. So the trade-off is worthy. Meanwhile, this also hints the important effect of neighborhood reversible flow on temporal stability or consistency.

	\begin{figure}[!t]
		\centering {
			\includegraphics[width=1.0\columnwidth]{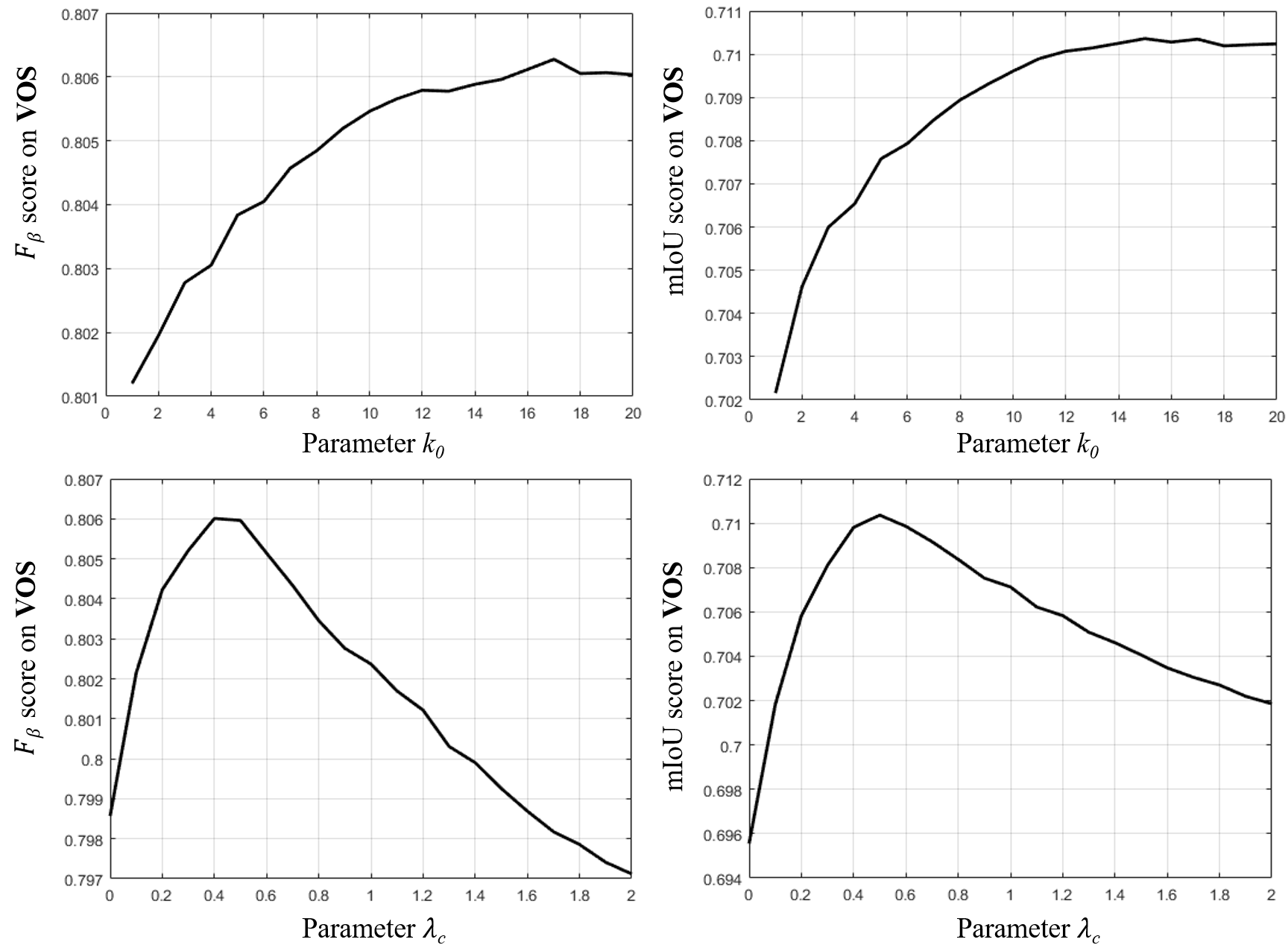}}
		\caption{Influence of parameters $k_0$ and $\lambda_c$ to our approach.}
		\label{Fig:parameter}
	\end{figure}
	
	\bl{Parameter setting}.~~In the experiment  based on CCNN, we smoothly vary two key parameters used in NRF, including the $k_0$ in constructing neighborhood flow and the $\lambda_c$ that controls the strength of temporal propagation. As shown in Fig.~\ref{Fig:parameter}, larger $k_0$ tends to bring slightly better performance, while our approach performs the best when $\lambda_c=0.5$. In experiments, we set $k_0=15$ and $\lambda_c=0.5$ in constructing the neighborhood reversible flow.

	\bl{Selection of color spaces}.~~In constructing the flow, we represent each superpixel with three color spaces. As shown in Table~\ref{tab:colorspace}, a single color space performs slightly worse than their combinations. Actually, using multiple color spaces have been proved to be useful in detecting salient objects~\cite{JiangWYWZL13}, while multiple color spaces make it possible to assess temporal correspondences from several perspectives with a small growth in time cost. Therefore, we choose to use RGB, Lab and HSV color spaces in characterizing a superpixel.

	\begin{table}[!t]
		\centering{
			\caption{Performance of our CCNN based approach on \bl{VOS} when using different color space in constructing neighborhood reversible flow.}
			\label{tab:colorspace}
			\begin{tabular}{ccccc}
				\toprule
				Color Space                   & mAP  & mAR  & $F_\beta$ & mIoU \tabularnewline
				\midrule
				RGB                           & .785 & .862 & .801 & .703  \tabularnewline
				Lab                           & .786 & .860 & .802 & .702  \tabularnewline
				HSV                           & .787 & .866 & .804 & .707  \tabularnewline
				RGB+Lab+HSV                   & .789 & .870 & .806 & .710  \tabularnewline
				\bottomrule
			\end{tabular}
		}
	\end{table}
	
	\subsubsection{Running Time}\label{sec:Analysis1 Time}
	
	\begin{table}[t]
		\centering{
			\caption{Speed of key steps in our approach. Mark + means using multi-test.} \label{tab:speed}
			\begin{tabular}{ccc}
				\toprule
				Key Step                                & Speed (s/frame) \tabularnewline
				\midrule
				Initialization(+)                          &~ ~0.05(0.36) \tabularnewline
				Superpixel \& Feature(+)                   &~ ~0.12(0.12) \tabularnewline
				Build Flow \& Propagation(+)               &~ ~0.02(0.26) \tabularnewline
				Primary Object Seg.(+)                     &~ ~0.01(0.01) \tabularnewline
				Total(+)                                   &$\sim$0.20(0.75) \tabularnewline
				\bottomrule
			\end{tabular}
		}
	\end{table}
	
	% {\color{red}your content}
	We test the speed of the proposed approach with a 3.4GHz CPU (only use single thread) and a NVIDIA TITAN Xp GPU (without batch processing). The average time cost of each key step of our approach in processing $400\times{}224$ frames are shown in Table~\ref{tab:speed}. Note that the majority of the implementation runs on the Matlab platform with several key steps written in C (\eg, superpixel segmentation and feature conversion between pixels and superpixels). We find that our approach takes only 0.20s to process a frame if not using multi-test, and no more than 0.75s even using, which is much faster than many video-based models (\eg, 19.0s for NLC, 6.1s for ACO, 5.8s for FST, 5.4s for SAG and 4.7s for GF). This may be caused by the fact that we only build correspondences on superpixels with the neighborhood reversibility, which is very efficient. Moreover, we avoid using complex optimization objectives and constraints. Instead, we use only simple quadratic optimization objectives so as to obtain analytic solutions. The high efficiency of our approach makes it possible to be used in some real-world applications.
	
	\subsubsection{Failure Cases}\label{sec:Analysis1 Failure}
	
\begin{figure}[!t]
		\centering {
			\includegraphics[width=1.0\columnwidth]{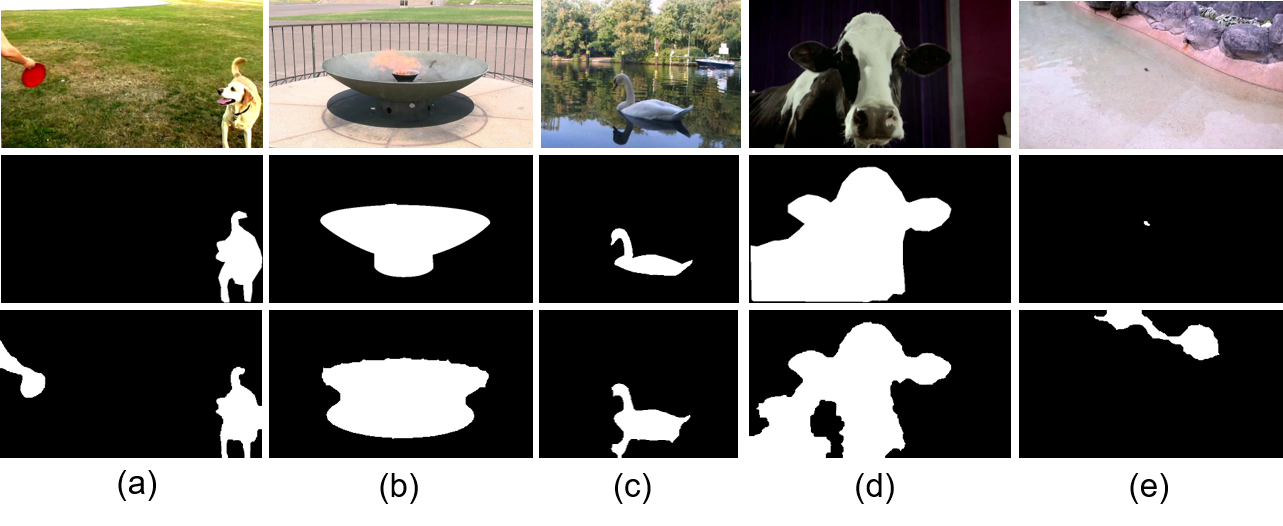}} %
		\caption{Failure cases of our approach. Rows from top to bottom: video frames, ground-truth masks and our results.}
		\label{Fig:failure}
	\end{figure}

	Beyond the successful cases, we also show in Fig.~\ref{Fig:failure} some failures. We find that failures can be caused by the way of defining primary objects. For example, the salient hand in Fig.~\ref{Fig:failure} (a) is not labeled as primary object as the corresponding videos from \bl{Youtube-Objects} are tagged with ``dog''. Moreover, shadow (Fig.~\ref{Fig:failure} (b)) and reflection (Fig.~\ref{Fig:failure} (c)), generated by the target object and environment, are some other reasons that may cause unexpected failures due to their similar saliency with the target object. It is also easily to fail when part regions of the target salient object are similar to background (Fig.~\ref{Fig:failure} (d)). Specially, successful segmentation is also very hard for some minuscule objects, \eg, the crab in water (Fig.~\ref{Fig:failure} (e)). Such failures need further exploration in future.
	
	\section{Conclusion}\label{sec:Conclusion}
	
	In this paper, we propose a simple yet effective approach for primary video object segmentation. Based on the complementary relationship of foreground and background, the problem of primary object segmentation is turned into an optimization problem of objective function. According to the proposed objective function, a complementary convolutional neural network is designed and trained on massive images from salient object datasets to handle complementary tasks. Then by the trained models, the foreground and background in a video frame can be effectively predicted from the spatial perspective. After that, such spatial predictions are efficiently propagated via the inter-frame flow that has the characteristic of neighborhood reversibility. In this manner, primary objects in different frames can gradually pop-out, while various types of distractors can be well suppressed. Extensive experiments on three video datasets have validated the effectiveness of the proposed approach.
	
	In the future work, we tend to improve the proposed approach by fusing multiple ways of defining primary video objects like motion patterns, semantic tags and human visual attention. Moreover, we will try to develop a completely end-to-end spatiotemporal model for primary video object segmentation by incorporating the recursive mechanism.

	\ifCLASSOPTIONcompsoc
%	\section*{Acknowledgments}
%	\else
%	\section*{Acknowledgment}
%	\fi
%	This work was partially supported by National Natural Science Foundation of China (61672072 and 61532003) and Beijing Science and Technology Nova Program (Z181100006218063). Jia Li is the corresponding author. 
	
	\bibliographystyle{IEEEtran}
	\bibliography{bibNRFPlus}
	
		\begin{IEEEbiography}[{\includegraphics[width=1in,height=1.25in,clip,keepaspectratio]{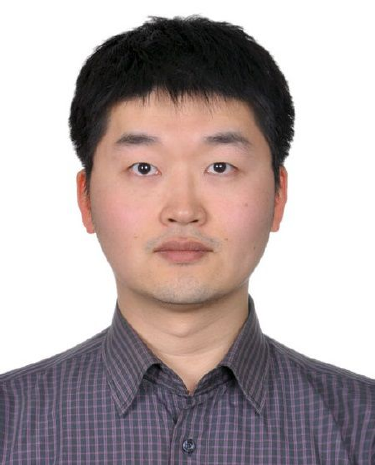}}]{Jia Li}
			is currently an associate professor with the State Key Laboratory of Virtual Reality Technology and Systems, School of Computer Science and Engineering, Beihang University. He received his B.E. degree from Tsinghua University in 2005 an Ph.D. degree from the Institute of Computing Technology, Chinese Academy of Sciences in 2011. His research interests include computer vision and image/video processing. He is a senior member of IEEE.
		\end{IEEEbiography}
	
		\begin{IEEEbiography}[{\includegraphics[width=1in,height=1.25in,clip,keepaspectratio]{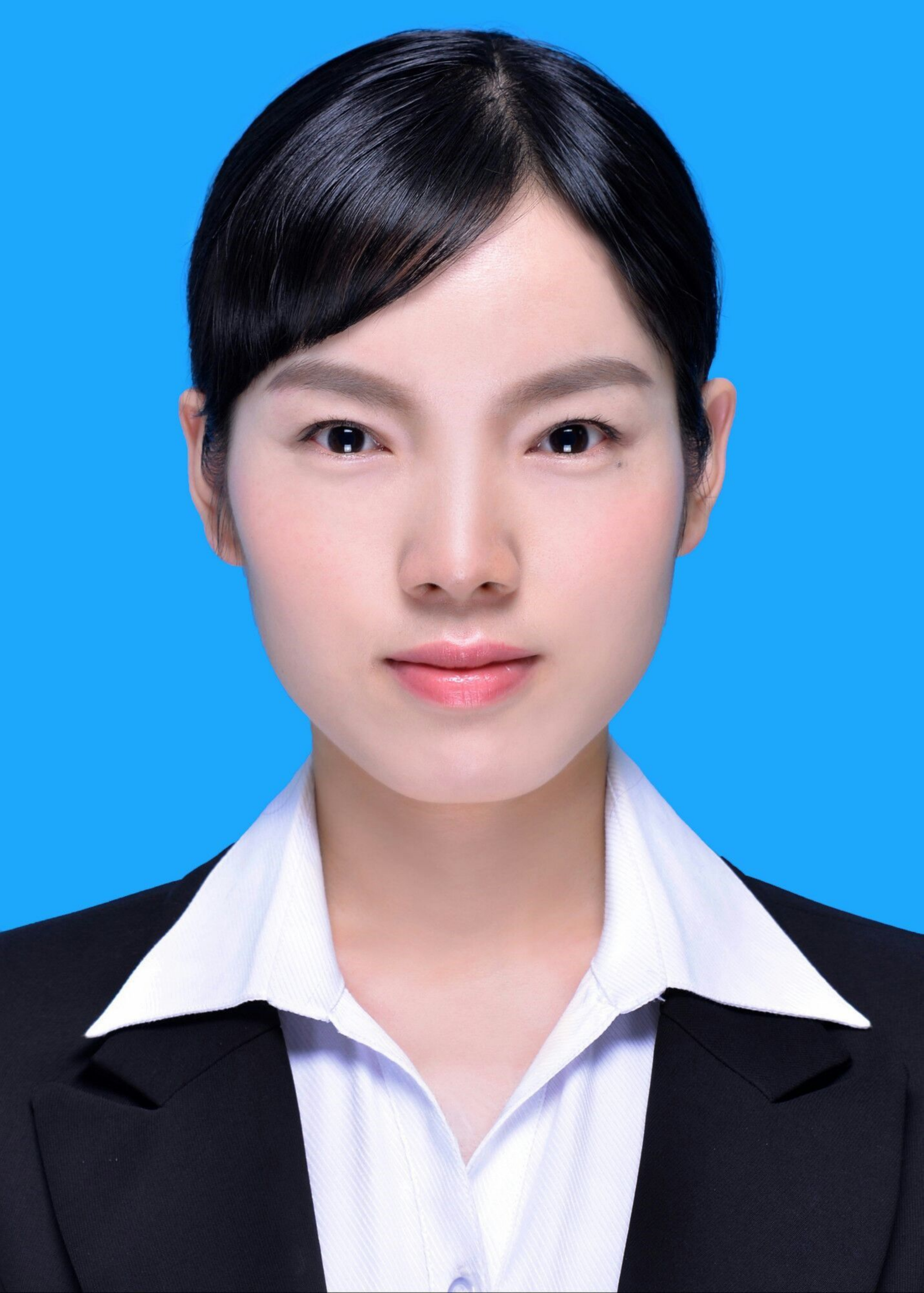}}]{Junjie Wu}			
			is currently working towards the PhD degree with the State Key Laboratory of Virtual Reality Technology and Systems, School of Computer Science and Engineering, Beihang University. Her research interests include computer vision and image/video processing. 			
		\end{IEEEbiography}
	
		\begin{IEEEbiography}[{\includegraphics[width=1in,height=1.25in,clip,keepaspectratio]{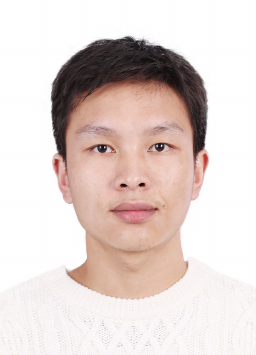}}]{Anlin Zheng} is currently a researcher with the Megvii Inc. (Face++). He received the MS degree from the State Key Laboratory of Virtual Reality Technology and Systems, School of Computer Science and Engineering, Beihang University in 2018. His research interests include computer vision and deep learning.	
		\end{IEEEbiography}
	
		\begin{IEEEbiography}[{\includegraphics[width=1in,height=1.25in,clip,keepaspectratio]{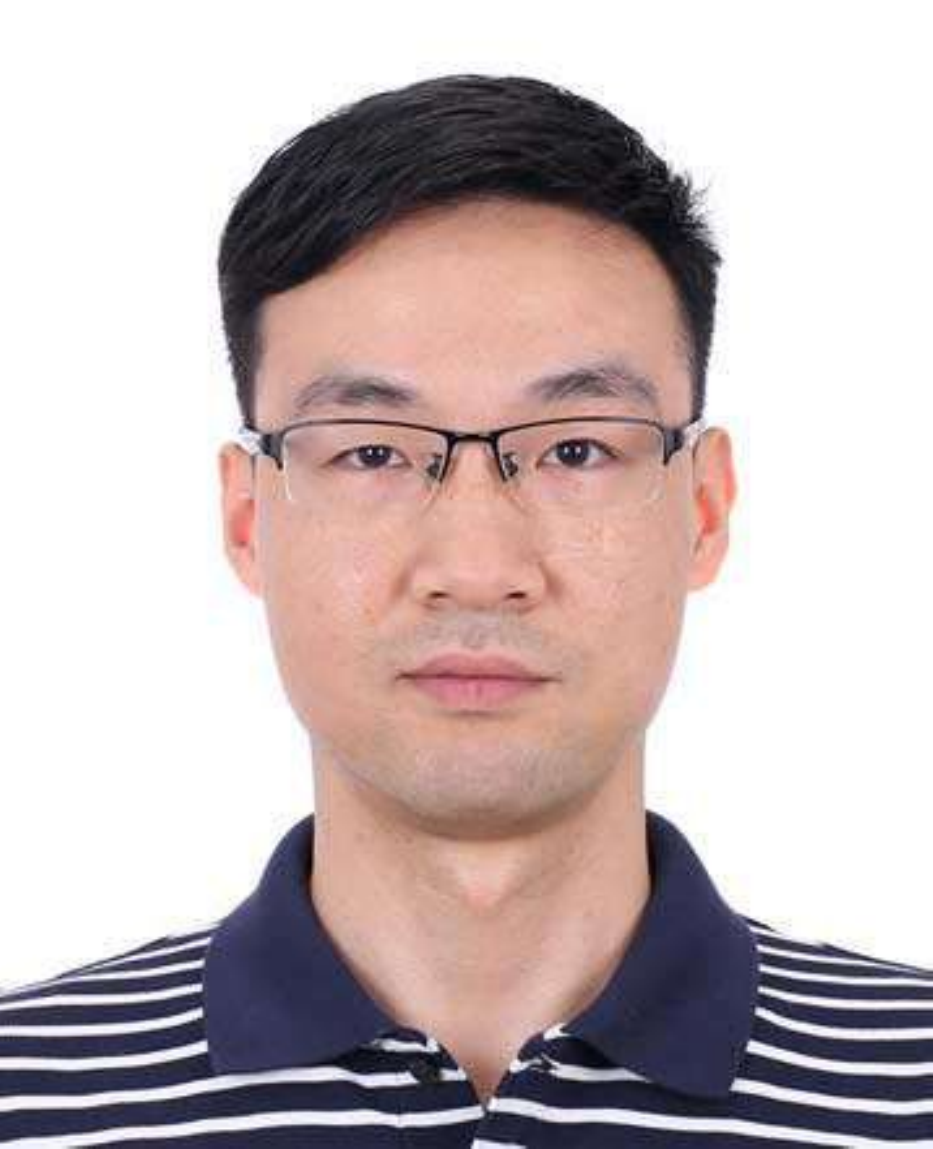}}]{Yafei Song}
			is currently a postdoctoral researcher with the School of Electronics Engineering and Computer Science, Peking University, Beijing, China, and the A.I. Labs, Alibaba Group, Beijing, China. He received the Ph.D. degree from the State Key Laboratory of Virtual Reality Technology and Systems, School of Computer Science and Engineering, Beihang University in 2017. His research interests include computer vision, machine learning, augmented reality, and robotics.			
		\end{IEEEbiography}
	
		\begin{IEEEbiography}[{\includegraphics[width=1in,height=1.25in,clip,keepaspectratio]{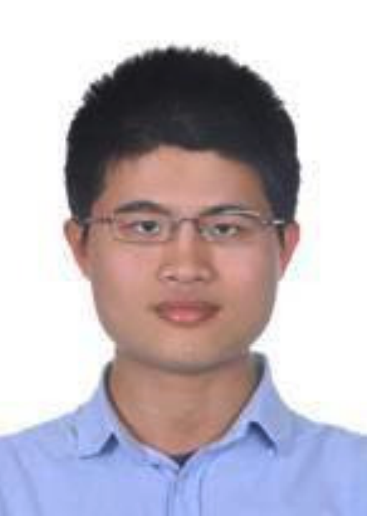}}]{Yu Zhang}
			is currently a researcher with SenseTime Research. He received the Ph.D. degree from the State Key Laboratory of Virtual Reality Technology and Systems, School of Computer Science and Engineering, Beihang University in 2018. His research interests include computer vision and deep learning.			
	    \end{IEEEbiography}
		
		\begin{IEEEbiography}[{\includegraphics[width=1in,height=1.25in,clip,keepaspectratio]{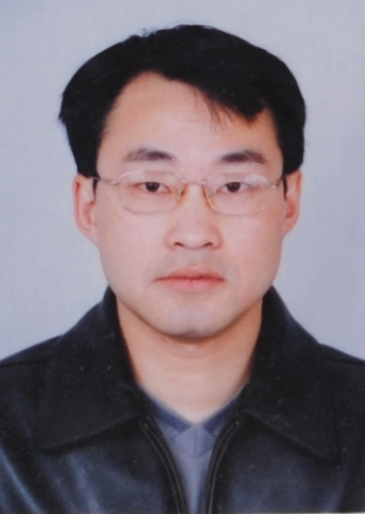}}]{Xiaowu Chen} is with the State Key Laboratory of Virtual Reality Technology and Systems, School of Computer Science and Engineering, Beihang University. He received his Ph.D. degree in computer science from Beihang University in 2001. His research interests include virtual reality, augmented reality, computer graphics and computer vision. He is a senior member of IEEE.
		\end{IEEEbiography}
		
		%\newpage

		
		%\vfill
		

\end{document}